\documentclass[journal, twoside, final]{IEEEtran}
\usepackage{amsmath}
\usepackage{amsfonts}
\usepackage{sansmath}
\usepackage{graphicx}
\usepackage{caption}
\usepackage{rotating} 
\usepackage{color, colortbl}	
\usepackage{amsmath}
\usepackage{amssymb}
\usepackage{graphicx}
\usepackage{float}
\usepackage{url}
\usepackage{longtable}  
\usepackage{array}
\usepackage{multirow}
\usepackage[dvipsnames]{xcolor}
\usepackage{soul}
\usepackage[colorlinks]{hyperref}
\usepackage{graphicx}
\usepackage{subcaption}
\hypersetup{
    linkcolor=blue,
    filecolor=magenta,
    urlcolor=teal,
    citecolor=magenta
    }
\usepackage[noadjust]{cite} 
\usepackage{graphicx} % for figures
\usepackage{lscape}
\usepackage{bm} % for bold symbols in equations
\usepackage{amssymb}

\usepackage{array,multirow} % for the table that uses multirows
\graphicspath{ {./images/} }

\usepackage{stackengine}

\def\BibTeX{{\rm B\kern-.05em{\sc i\kern-.025em b}\kern-.08em
    T\kern-.1667em\lower.7ex\hbox{E}\kern-.125emX}}
    
\begin{document}
%\bstctlcite{IEEEexample:BSTcontrol} %% Added to shorten the author lists and replace by et al 

% \title{Multimodal Cardiac Waveform Learning}

% \title{Data-Driven Multimodal Learning for Cardiac Waveform Reconstruction and Prediction}

% \title{Data-Driven Multimodal Learning for Cardiac Waveform Reconstruction and Prediction}

% \title{Exploring Joint and Exclusive Cardiac Information Through Multimodal Waveform Learning}

% \title{Multimodal Learning of Cardiac Waveforms: Learning Joint and Exclusive Information}

% \title{Enhanced Cardiac Monitoring through Multimodal Learning: Integrating ECG with Complementary Modalities}

% \title{Towards Multimodal Cardiac Monitoring: Integrating ECG with Complementary Modalities for Enhanced Cardiac Monitoring}

% \title{Integrating ECG with Complementary Modalities: Exploring Common and Exclusive Characteristics of Multimodal Cardiac Waveforms}

\title{Cross-Learning Between ECG and PCG: Exploring Common and Exclusive Characteristics of Bimodal Electromechanical Cardiac Waveforms}

%\title{Electro-mechanical Modulation of Heart: Analysis of Linear and Nonlinear Causality (or Information Flow) for Simultaneous ECG and PCG Signal}

\author{Sajjad~Karimi, Amit~J~Shah, Gari~D.~Clifford~\IEEEmembership{Fellow,~IEEE}, and Reza~Sameni\textsuperscript{*} \IEEEmembership{Senior~Member,~IEEE}%
\thanks{The authors are with the School of Medicine, Emory University. A.J. Shah is also with the Rollins School of Public Health, Department of Epidemiology, Emory University. G.D.~Clifford and R.~Sameni are also with the Department of Biomedical Engineering, Georgia Institute of Technology and Emory University (email: \url{rsameni@dbmi.emory.edu}).
}%
}
%//////////////////////////////////////////
% \author{
% \thanks{Manuscript received SEP DD, 2024; revised YY, 2024.}%
% \thanks{Copyright (c) 2024 IEEE. Personal use of this material is permitted.  However, permission to use this material for any other purpose must be obtained from the IEEE by emailing pubs-permissions@ieee.org.}
% % \newline
% \thanks{The other authors are with the Department of Biomedical Informatics, Emory University, Atlanta, GA, USA. R. Sameni are also with the Biomedical Engineering Department, Georgia Institute of Technology. \href{rsameni@dbmi.emory.edu}{rsameni@dbmi.emory.edu}}
% }
%//////////////////////////////////////////
\markboth{}{}
\maketitle
%//////////////////////////////////////////
% future ideas are seq2seq transformers & CycleGan for both translation (registration) and sequence forecasting, Granger Causality in both single and multi-modal inputs
% Electromechanical Synchronization During Stress Test

\begin{abstract}

Simultaneous electrocardiography (ECG) and phonocardiogram (PCG) provide a comprehensive, multimodal perspective on cardiac function by capturing the heart’s electrical and mechanical activities, respectively. However, the distinct and overlapping information content of these signals, as well as their potential for mutual reconstruction and biomarker extraction, remains incompletely understood, especially under varying physiological conditions and across individuals. 

In this study, we systematically investigate the common and exclusive characteristics of ECG and PCG using the EPHNOGRAM dataset of simultaneous ECG-PCG recordings during rest and exercise. We employ a suite of linear and nonlinear machine learning models, including non-causal LSTM networks, to reconstruct each modality from the other and analyze the influence of causality, physiological state, and cross-subject variability. Our results demonstrate that nonlinear models, particularly non-causal LSTM, provide superior reconstruction performance, with reconstructing ECG from PCG proving more tractable than the reverse. Exercise and cross-subject scenarios present significant challenges, but envelope-based modeling that utilizes instantaneous amplitude features substantially improves cross-subject generalizability for cross-modal learning. Furthermore, we demonstrate that clinically relevant ECG biomarkers, such as fiducial points and QT intervals, can be estimated from PCG in cross-subject settings. 
These findings advance our understanding of the relationship between electromechanical cardiac modalities, in terms of both waveform characteristics and the timing of cardiac events, with potential applications in novel multimodal cardiac monitoring technologies.

% Conclusion:
% PCG to ECG fiducial 
\end{abstract}
%////////////////////////////////

\section{Introduction}

Cardiac function is inherently multimodal, arising from tightly coupled electrical, mechanical, hemodynamic, autonomic, and metabolic processes that generate distinct but interrelated biosignals. Specifically, the electrocardiogram (ECG) records the heart's electrical activity, providing insight into conduction and repolarization mechanisms, while cardiac auscultation---one of the oldest and most fundamental clinical tools---captures heart sounds produced by valve motion and blood flow. These sounds are digitally recorded as phonocardiogram (PCG) sounds~\cite{liu2016open}. ECG and PCG are both accessible, low-cost, and complementary: ECG reflects the electrical function of the heart, and PCG reflects its mechanical function. Several studies have shown that their integration offers a more comprehensive approach to cardiac monitoring and screening than either modality alone~\cite{karimi2024electromechanical,castro2015analysis, han2023multimodal}.

Recent advances in sensor technology and machine learning have accelerated the integration of ECG and PCG for cardiovascular monitoring~\cite{hettiarachchi2021novel}, disease detection~\cite{singhal2023cardiovascular}, and biomarker extraction~\cite{tripathi2022multilevel, huang2024deep, ajitkumar2021heart}. Hardware innovations, such as polymer-based dry electrode stethoscopes and 3D-printed acoustic devices, now enable simultaneous ECG and PCG acquisition~\cite{shi2024design, monteiro2023novel}. These developments, coupled with large, synchronized ECG-PCG datasets, have spurred new research directions. Multimodal frameworks have demonstrated improved performance over single-modal approaches in tasks such as arrhythmia classification~\cite{hangaragi2025integrated}, heart sound segmentation, and coronary artery disease detection~\cite{sun2024enhanced}. Recent studies have further highlighted the value of coupling signals~\cite{karimi2024electromechanical} and nonlinear feature extraction to capture the intrinsic relationship between cardiac electrical and mechanical activity, improving disease classification and risk assessment~\cite{singhal2023cardiovascular, sun2024enhanced}.

Despite these advances, several fundamental questions remain regarding the degree to which ECG and PCG share recoverable information, the nature of their exclusive components, and the feasibility of reconstructing one modality from the other, especially under different physiological states (rest vs.\ exercise) or across subjects. Prior work has shown that electrode and stethoscope placement, physiological variability, and noise substantially impact signal morphology and waveform modeling accuracy~\cite{scholzel2016can, clifford2006advanced}. Most studies have focused on classification or limited waveform translation, often neglecting the detailed evaluation of information flow, causality, and biomarker recovery. While some studies have attempted to reconstruct ECG from PCG using deep learning~\cite{chen2023deep}, their evaluations have often been limited to simple error metrics, such as root mean square error (RMSE), neglecting the assessment of clinically relevant features and robustness against noise and physiological variation. Compounded by the fact that PCG does not reliably predict the amplitude or morphology of ECG waves, further challenges arise from variable electrode placement that can introduce large qualitative differences in ECG shape~\cite{scholzel2016can}. Other knowledge gaps in this area include: direction of information flow (in a causal sense), between multimodal cardiac waveforms and the corresponding timing of cardiac events and biomarkers such as the QT interval or QRS width from the ECG versus dominant PCG-based events such as the S1 and S2 waves.

The applications of bimodal ECG-PCG modeling and processing are manifold, including: (1) bimodal cardiac monitoring and improving signal denoising and reconstruction by leveraging the complementary strengths of ECG and PCG, particularly for recovering poor-quality segments in wearable cardiac monitors; (2) enabling bimodal biomarker extraction by utilizing both shared and unique physiological information present across the two modalities; (3) studying the causal/non-causal \textit{flow of information} between ECG and PCG to better understand the interdependence of electrical and mechanical cardiac activity; (4) inferring systolic and diastolic timing in PCG from ECG, and estimating ECG parameters (e.g., QT interval) from PCG alone; and (5) identifying modality-specific innovations ~\cite{phanphaisarn2011heart}, by analyzing residuals---i.e., the components of a signal that cannot be reconstructed from the other modality---which can serve as valuable features for bimodal AI/ML model development.

This work investigates the causal and non-causal relationships between ECG and PCG using simultaneous recordings during rest and exercise from the EPHNOGRAM dataset~\cite{Kazemnejad2024, EPHNOGRAMDataset}. We evaluate cross-modal signal reconstruction and event timing prediction to quantify information transfer between electrical and mechanical cardiac signals under varying physiological states and across individuals. The analysis focuses on both waveform reconstruction and the accuracy of extracting clinically relevant biomarkers, such as QRS duration and QT interval, when estimating ECG from PCG and vice versa. Results characterize the potential and limitations of cross-modal modeling for robust and generalizable cardiac monitoring applications.

Through the proposed unified framework in the current study, we advance the understanding of multimodal cardiac signal integration and establish robust, generalizable methods for waveform reconstruction and biomarker extraction. Our findings have direct implications for the development of wearable and remote cardiac monitoring technologies, particularly in resource-limited or ambulatory settings~\cite{monteiro2023novel, sun2024enhanced, hangaragi2025integrated}.

The remainder of the paper is organized as follows: Section~\ref{sec:methodology} details the EPHNOGRAM dataset and preprocessing steps, describing the modeling frameworks and evaluation criteria. Section~\ref{sec:result} presents the main results, including waveform reconstruction, spectrum analysis, and biomarker recovery. Section~\ref{sec:discussion} discusses the implications, limitations, and future directions, while Section~\ref{sec:conclusion} concludes the work.

\section{Methodology}
\label{sec:methodology}

%/////////////////////////////////////////
\subsection{The EPHNOGRAM simultaneous ECG-PCG dataset}
\label{sec:dataset}

The EPHNOGRAM dataset \cite{Kazemnejad2024, EPHNOGRAMDataset}, publicly available on PhysioNet, contains 68 simultaneous ECG-PCG recordings collected from 24 healthy participants (age 25.4$\pm$1.9 years) during physical activities. Data were acquired using single-channel ECG and PCG stethoscopes in an indoor fitness center, including 8 short recordings of 30 seconds each and 60 long recordings of 30 minutes each. Activities include:

\begin{enumerate}
    \item \textit{Scenario A --- Resting:} 30-minute supine recordings and short seated samples in a quiet room.
    \item \textit{Scenario B --- Walking:} Treadmill walking at 3.7\,km/h.
    \item \textit{Scenario C --- Treadmill stress-test:} Modified Bruce protocol \cite{bruce1971exercise}, with progressive intensity until termination criteria (fatigue, excessive heart rate, or chest pain), followed by cooldown.
    \item \textit{Scenario D --- Bicycle stress-test:} Progressive workload (25\,W/min increase) on a stationary bike until termination criteria, followed by gradual workload reduction.
\end{enumerate}

All records are available in WFDB format and activity intensity time labels. The full data collection protocol is detailed in \cite{Kazemnejad2024,EPHNOGRAMDataset}.

In this study, we focus on 28 recordings from EPHNOGRAM: 9 from rest and walking (Scenario A-B) and 19 from stress tests (Scenarios C-D). Records with excessive noise or electrode disconnections (as noted in \texttt{ECGPCGSpreadsheet.csv} signal quality notes of EPHNOGRAM \cite{EPHNOGRAMDataset}) were excluded. The included recordings met cleanliness criteria, avoided signal saturation, and completed the full test protocol \cite{Kazemnejad2024}. 

The selected files follow the naming convention \texttt{ECGPCG00XY.mat}: Rest recordings include \texttt{XY} values of 10, 11, 13, 14, 15, 16, 21, 22, and 23. For stress tests, \texttt{XY} values are 01, 25, 27, 29, 30, 32, 33, 34, 36, 38, 47, 52, 55, 61, 62, 64, 66, 67, and 68.

%////////////////////////////////////////////////////////////////////////

\subsection{Preprocessing}
ECG baseline wander was corrected using a 0.2--30\,Hz bandpass filter. Power-line interference (50\,Hz) was attenuated via a second-order IIR notch filter (Q-factor = 45) with zero-phase forward-backward implementation. 

PCG signals, which represent transient acoustic vibrations centered around zero and lacking intrinsic baseline drift and power-line, were bandpass filtered (10--200\,Hz). 
For both modalities, outliers exceeding $\pm6\sigma(t)$ were clipped, where $\sigma(t)$ denotes the time-varying standard deviation computed over 1-minute sliding windows across each 30-minute recording. 
Based on an expert review of the EPHNOGRAM dataset, this threshold was chosen to optimally balance outlier removal and feature preservation, like the R-peak. Lower thresholds (e.g., $3\sigma$) distorted the QRS complex, while higher thresholds (e.g., $9\sigma$ or $12\sigma$) failed to eliminate outliers. This choice is dataset-specific and was determined empirically rather than as a general rule.

Signal normalization was conducted using a time-adaptive approach to address amplitude variability during the stress test, consistent with established literature. Since the differences in overall signal intensity cannot be directly interpreted from the other modalities, this normalization strategy enhances cross-subject modeling by diminishing inter-subject amplitude discrepancies. Specifically, each signal $x(t)$ underwent normalization as follows:
\begin{equation}
    z(t) = \frac{x(t) - m(t)}{\sigma(t)}
\end{equation}
where $m(t)$ and $\sigma(t)$ are the moving average and standard deviation calculated over 60-second windows sliding through the entire recording. Due to the baseline correction, $m(t)$ is close to zero for both ECG and PCG. This addresses the amplitude variability inherent in exercise stress tests~\cite{scholzel2016can}, which is particularly critical for long-duration signals. The time-varying normalization ensures local signal characteristics are preserved while mitigating global amplitude discrepancies across subjects and recording sessions, enabling robust cross-subject model training. Processed signals serve as inputs for machine learning and reconstruction analyses.
Finally, both signals were resampled from 8000\,Hz to 1000\,Hz to reduce computational costs and memory usage in future analyses.

\subsection{Multimodal Waveform Learning Frameworks}
\label{sec:method}

\begin{figure}[tb]
    \includegraphics[trim={0cm 0cm 0cm 0cm},clip,width=0.8\linewidth]{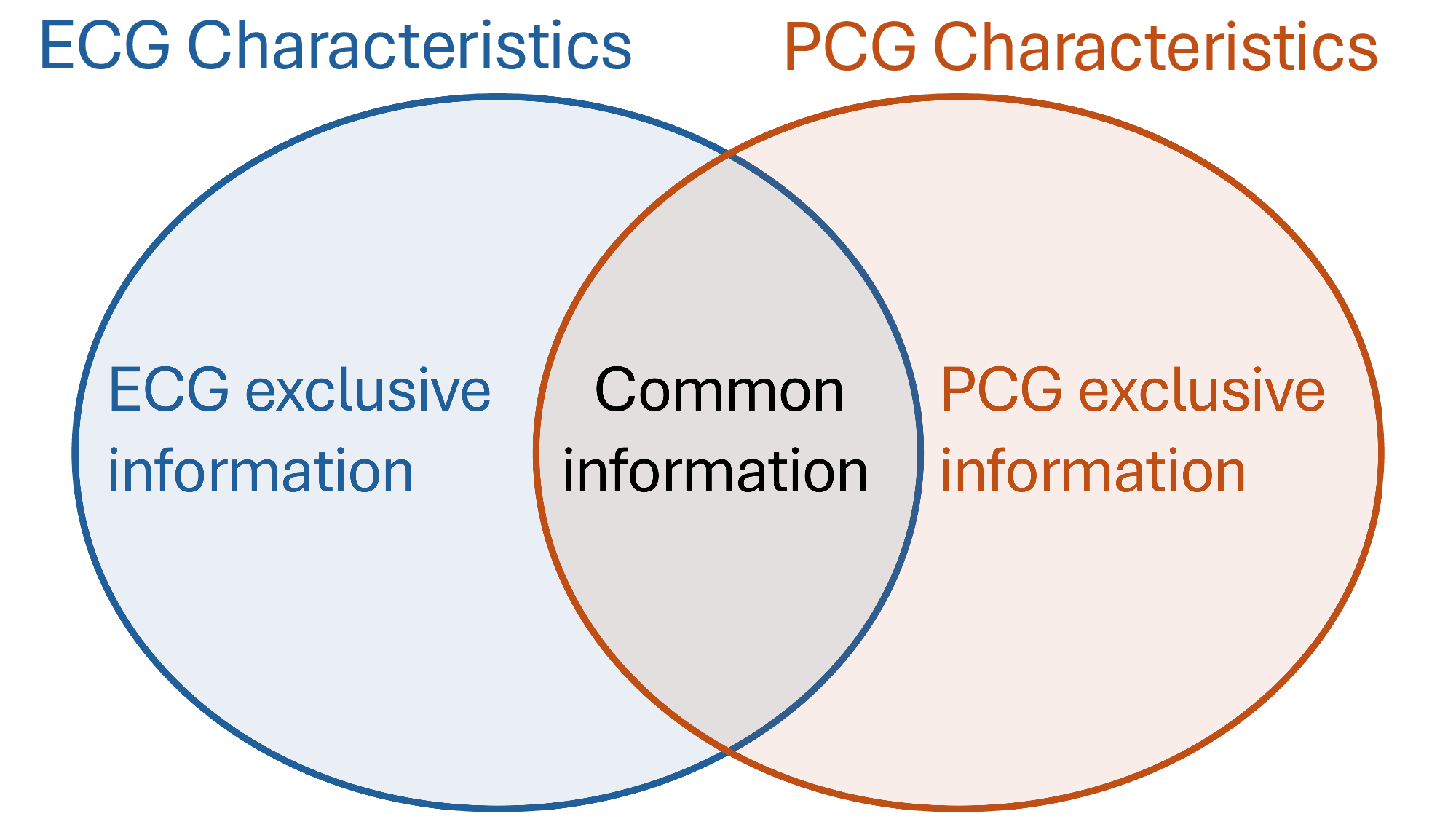}
    \caption{Venn diagram illustrating the information structure of ECG and PCG in representing cardiac activity. While rhythms are shared between both modalities, each also has exclusive characteristics that are complementary.}
    \label{fig:venn-diag}
\end{figure}

ECG and PCG signals offer complementary insights into cardiac function. Specifically, ECG captures the electrical activity of the heart, while PCG reflects the mechanical events associated with cardiac contraction and relaxation. While these modalities share certain physiological information, each contains unique and exclusive characteristics specific to the underlying cardiac processes. Figure~\ref{fig:venn-diag} illustrates this relationship through a Venn diagram, highlighting both the shared and exclusive components represented by ECG and PCG.

% This conceptual framework is further clarified in Fig.~\ref{fig:nspectra}, which displays the normalized power spectral densities (PSDs) of ECG and PCG signals from the EPHNOGRAM dataset, utilizing the Welch method over 1-second windows with a resolution of 1\,Hz and 50\% overlap between segments. The PSD analysis demonstrates that low-frequency components (below 5\,Hz) are predominantly unique to ECG, whereas high-frequency components (above 50\,Hz) are characteristic of PCG. The intermediate frequency range (5--50\,Hz) likely contains information common to both modalities, reflecting shared physiological processes.

\begin{figure}[tb]
    \centering
    \includegraphics[trim={0cm 0cm 0cm 0cm},clip, width=0.8\columnwidth]{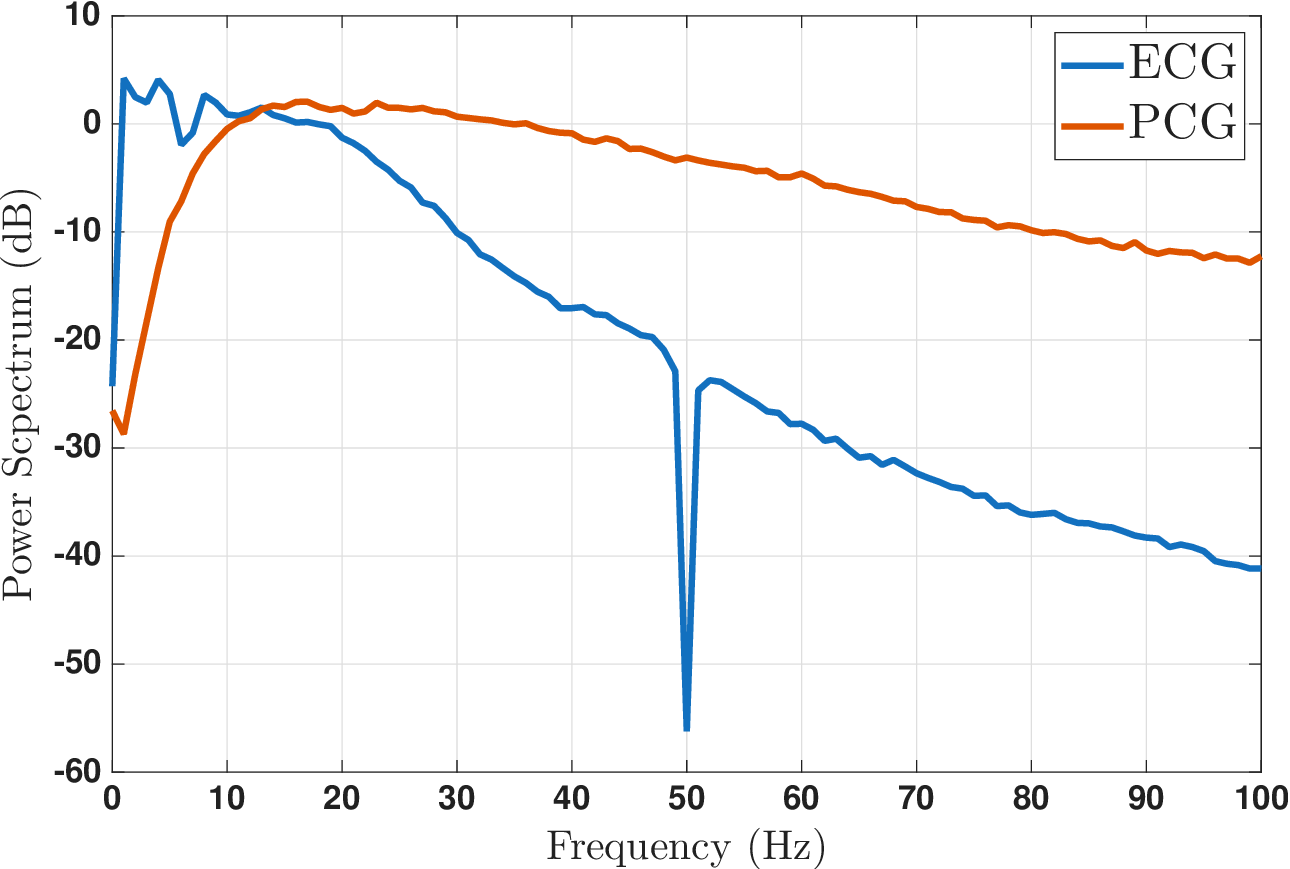}
    \caption{Power spectrum for ECG and PCG across all recordings in EPHNOGRAM dataset.}
    \label{fig:nspectra}
\end{figure}

This conceptual framework is further clarified in Fig.~\ref{fig:nspectra}, which displays the power spectra of ECG and PCG signals from the EPHNOGRAM dataset, utilizing the Welch method over 1-second windows with a resolution of 1\,Hz and 50\% overlap between segments. The power spectrum analysis demonstrates that low-frequency components (below 5\,Hz) are predominantly unique to ECG, whereas high-frequency components (above 50\,Hz) are characteristic of PCG. The intermediate frequency range (5--50\,Hz) likely contains information common to both modalities, reflecting shared physiological processes.

The subsequent sections will outline the methodological approach employed to differentiate between shared and exclusive characteristics of ECG and PCG, as well as the evaluation criteria adopted for the analysis. Particular attention will be given to the potential influence of noise in the reference signals, especially in light of the challenges associated with data acquisition during exercise stress testing.
 
\subsubsection{Machine learning frameworks}
\begin{figure}[tb]
    \centering
    \includegraphics[trim={0cm 0cm 0cm 0cm},clip,width=\linewidth]{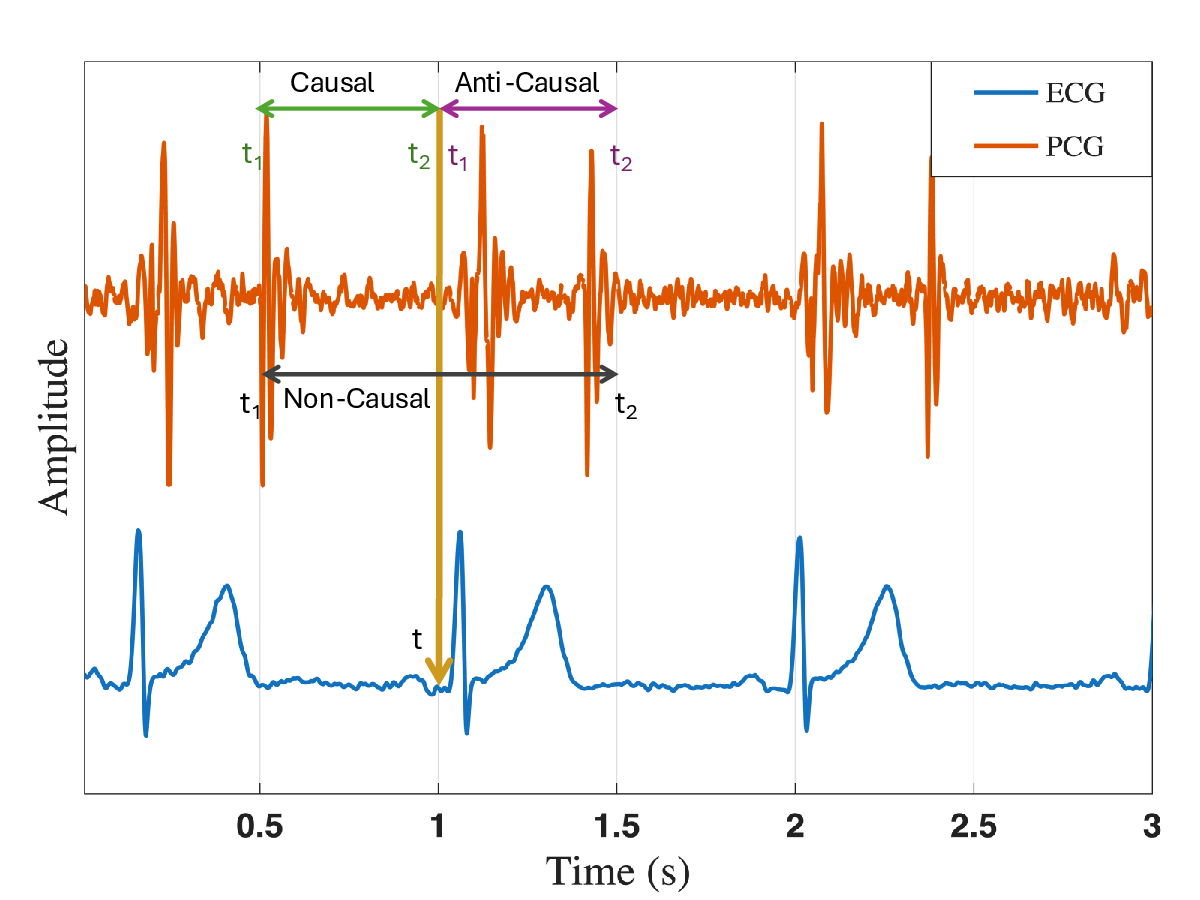}
    \caption{Illustration of three temporal learning frameworks for PCG$\rightarrow$ECG transformation at time $t=1$ and segment size $\Delta t=0.5$\,s: (a) causal (green segment $(0.5 \leq \tau \leq 1)$), utilizing only past and present input; (b) anti-causal (purple segment $(1 \leq \tau \leq 1.5)$), utilizing only future input; and (c) non-causal (black segment $(0.5 \leq \tau \leq 1.5)$), utilizing both past and future input within a symmetric window.}
    \label{fig:causality}
\end{figure}

Considering two simultaneously modalities, $x(t)$ and $y(t)$, we use $x \rightarrow y$ to denote models that learn $y(t)$ as a time series—or its characteristics--using $x(t)$ as input. Using this notation ECG$\rightarrow$PCG denotes predicting the PCG waveform/features from the ECG. Similarly, throughout, PCG$\rightarrow$ECG denotes learning the ECG waveform/features from the PCG. Taking the temporal aspect into account, this corresponds to:
\begin{equation}
    y(t) = g\left(x(\tau)\,|\, t_1 \leq \tau \leq t_2\right)
\end{equation}
where $t$ is the predicted time point, $[t_1, t_2]$ is the time interval of the input modality with duration $\Delta t = t_2-t_1$ used for prediction, and $g(\cdot)$ is the machine-learning model, which predicts $y$ from $x$. To study inter-modal causality, we consider three temporal schemes, as illustrated in Fig.~\ref{fig:causality}:

% This study examines the dynamic interdependence between cardiac intervals by estimating ECG and PCG signals from one another using three machine learning regression models. Each model is evaluated under three temporal frameworks: causal, anti-causal, and non-causal input features, as illustrated in Fig.~\ref{fig:causality}.
% \begin{itemize}
%     \item \textbf{Causal:} In the causal scenario, the model uses input segments from $\Delta t$ before time $t$ up to $t$ to estimate the target modality at time $t$. This approach ensures that only present and past information is utilized, aligning with real-time, physically realizable systems.
%     \item \textbf{Anti-causal:} The anti-causal scenario uses input segments from $t$ to $\Delta t$ after $t$ to estimate the target at time $t$. Here, only future information is employed, which is not physically realizable in real-time but is informative for analyzing the directionality of information flow.
%     \item \textbf{Non-causal:} The non-causal scenario uses a symmetric segment from $\Delta t$ before to $\Delta t$ after $t$ as input for estimating the target at $t$. This framework leverages both past and future information, providing an upper bound on achievable reconstruction performance.
% \end{itemize}

\begin{itemize}
    \item \textit{Causal ($t_2 \leq t$)} Uses input segments of duration $\Delta t$ immediately before time $t$ (i.e., $[t-\Delta t,\, t]$) to estimate the target at $t$; only past and present information is used.
    \item \textit{Anti-causal ($t < t_1$):} Uses input segments of duration $\Delta t$ immediately after time $t$ (i.e., $[t,\, t+\Delta t]$) to estimate the target at $t$; only future information is used.
    \item \textit{Non-causal ($t_1 < t < t_2$):} Uses a segment spanning before and after $t$ from one modality to predict the other. While this segment can be generally asymmetric around $t$, for the current study, we use a symmetric segment centered around $t$.
\end{itemize}

% Three distinct regression models were selected to comprehensively capture the spectrum of possible relationships between ECG and PCG signals, ranging from simple linear associations to complex nonlinear and temporal dynamics. 

We consider three machine-learning models $g(\cdot)$ to investigate the potential relationships between ECG and PCG, ranging from basic linear patterns to nonlinear temporal dynamics:
\begin{itemize}
    \item \textit{Dynamic Linear Model (LASSO):} Utilizes a segment around time $t$ from one modality as input to estimate the target modality at $t$ using a linear combination of samples from the segment. This model captures linear dependencies between ECG and PCG. The current study employs LASSO algorithms to achieve robust estimation of combination coefficients~\cite{tibshirani1996regression}.
    
    \item \textit{Dynamic Nonlinear Neural Network (NN):} A two-layer fully connected neural network with rectified linear unit (ReLU) activations. The first layer contains 50 neurons with a clipped ReLU, the second layer has 25 neurons with a leaky ReLU, and a dropout layer (10\%) is included to mitigate overfitting. The feature and target definitions mirror those of the linear model.
    
    \item \textit{Long Short-Term Memory (LSTM) Network:} A nonlinear dynamic model with two LSTM layers (200 and 100 neurons) and leaky ReLU activations, followed by a dropout layer (10\%) and a final linear layer with 25 neurons to map the LSTM output to the target modality. The input and target definitions are consistent with those of the other models.
\end{itemize}

The selected NN and LSTM architectures were chosen to ensure sufficient model capacity for capturing the complex, nonlinear, and temporal patterns present in ECG and PCG signals. The network sizes and dropout rates were verified through pilot experiments to provide robust performance across both within-subject and cross-subject scenarios, while avoiding extensive hyperparameter tuning and ensuring a consistent structure for all case studies. This approach strikes a balance between modeling flexibility and computational efficiency, supporting fair comparisons across different experimental settings.

Together, these models represent a systematic progression from linear and interpretable to highly nonlinear and memory-based approaches, enabling a thorough evaluation of the modeling requirements for accurate ECG-PCG signal reconstruction.

\subsubsection{Training and testing protocols}

Both within-subject and cross-subject modeling approaches are examined for training and testing. Within-subject schemes account for intra-individual physiological variability, such as fatigue, while cross-subject models are trained on all subjects except one and tested on the held-out subject. Although cross-subject generalization is desirable for real-world applications, differences in cardiac dynamics and sensor placement can reduce performance on unseen subjects due to increased variability.

Thus, the following two validation schemes were implemented:
\begin{itemize}
    \item \textit{Within-subject:} 10-fold cross-validation was conducted on non-overlapping 3-minute segments from each subject's 30-minute recording (i.e., 10 segments). This ensures training and testing data come from the same individual but from different periods.
    \item \textit{Cross-subject:} Leave-one-out cross-validation (LOOCV) across the 28 selected recordings. Models were trained on 27 recordings and tested on the held-out recording.
\end{itemize}

To avoid transient effects at the beginning and end of each 30-minute recording, we excluded the first and last 1-second intervals from our calculations of different evaluation metrics. This approach enables us to thoroughly evaluate how well ECG and PCG information is transferred under different physiological conditions.
Since the number of training samples for the cross-subject scheme is nearly 30 times greater than that in the within-subject scheme, we considered double the number of parameters for NN and LSTM in the cross-subject scheme.

\subsection{Evaluation criteria}

% \subsubsection{Signal-to-noise ratio}
% Signal-to-noise ratio (SNR) quantifies the proportion of signal power to noise power in the reconstructed signal. It is commonly expressed in decibels (dB) as:
% \begin{equation}
%     \text{SNR}_{\mathrm{dB}} = 10 \log_{10} \left( \frac{\text{Signal Power}}{\text{Noise Power}} \right)
% \end{equation}
% Higher SNR values indicate better reconstruction quality. However, in this study, the true signal may contain noise due to simultaneous ECG and PCG recordings during stress tests. Therefore, SNR may not fully reflect reconstruction fidelity, and metrics such as correlation coefficient and coherence provide more robust assessments.

% \subsubsection{Cross correlation}
% Correlation Coefficient (CC)\\
% This measures the similarity between the original and reconstructed signals in terms of their linear relationship.\\
% Suitable when the goal is to maintain the overall pattern or trend of the signal rather than focusing on exact amplitude differences.

% \begin{equation}
%     \text{CC} = \frac{\sum_{i=1}^{N} (x_i - \bar{x})(\hat{x}_i - \bar{\hat{x}})}{\sqrt{\sum_{i=1}^{N} (x_i - \bar{x})^2 \cdot \sum_{i=1}^{N} (\hat{x}_i - \bar{\hat{x}})^2}}
% \end{equation}

\subsubsection{Signal-to-noise ratio}

The signal-to-noise ratio (SNR) is used to assess how well the reconstructed signal aligns with the original signal. In this context, the “signal” refers to the original ECG or PCG waveform, while the “noise” is defined as the difference between the original and reconstructed signals. A higher SNR indicates that the reconstruction closely follows the original waveform, with less residual error. If a model fails to identify patterns for reconstructing one modality from another, it may generate only zeros, resulting in an SNR of zero dB. However, it is important to note that, due to the nature of simultaneous ECG and PCG recordings—particularly during exercise—the original signals themselves may contain physiological noise and artifacts. As a result, SNR should be interpreted with caution and considered in conjunction with other evaluation measures.

\subsubsection{Cross correlation}

Cross-correlation, or the correlation coefficient, measures how closely the reconstructed signal follows the overall pattern or trend of the original signal, regardless of exact amplitude differences. This metric is particularly useful for assessing whether the main features and timing of the original waveform are preserved in the reconstruction, even if there are small differences in scale or offset.

\subsubsection{Cross coherence}
Cross coherence quantifies the frequency-specific linear relationship between the original and reconstructed signals \cite{bastos2016tutorial}. The coherence at each frequency $f$ is given by:

\begin{equation}
    \mu_{x\hat{x}}(f) = \frac{|P_{x\hat{x}}(f)|^2}{P_{xx}(f) P_{\hat{x}\hat{x}}(f)}
\end{equation}
where $P_{x\hat{x}}(f)$ denotes the cross-spectral density between the original signal $x(t)$ and the reconstructed signal $\hat{x}(t)$, and $P_{xx}(f)$ and $P_{\hat{x}\hat{x}}(f)$ represent the respective power spectral densities of $x(t)$ and $\hat{x}(t)$.
To summarize coherence across frequencies, the spectrum-weighted average coherence is used \cite{byeragani2003weighted, porges1980new}:

\begin{equation} 
    \bar{\mu}_{x\hat{x}} = \frac{\sum_{f} P_{xx}(f) \cdot \mu_{x\hat{x}}(f)}{\sum_{f} P_{xx}(f)}
\end{equation}

This approach emphasizes frequencies where the original signal has higher power, providing a more representative measure of coherence across the spectrum.

\subsubsection{ECG Fiducial Analysis and Biomarker Validation} 

The EPHNOGRAM dataset provides expert-annotated fiducial points for QRS complexes and T-waves in the ECG, enabling rigorous quantitative evaluation of reconstructed signals~\cite{kazemnejad2024open}. In this study, we applied the Latent Structure Influence Model (LSIM)-based fiducial detection algorithm from the Open-Source Electrophysiological Toolbox (OSET)~\cite{OSET3.14} to both original and reconstructed ECG signals. The LSIM algorithm identifies the onset and offset of QRS complexes and T-waves using the ECG waveform and detected R-peaks as inputs~\cite{karimi2020tractableinf, karimi2023tractablemle}. R-peaks were detected using a robust algorithm from OSET (\texttt{peak\_det\_likelihood\_long\_recs.m}). This approach enables the extraction of clinically relevant fiducial points, facilitating subsequent biomarker analysis, including assessment of the QT interval and QRS duration.

Detection performance was evaluated using mean absolute error (MAE) and root mean square error (RMSE). MAE measures the average absolute difference between the detected and expert-annotated fiducial points, providing a direct indication of typical timing error. RMSE reflects the square root of the average squared differences, placing greater emphasis on larger errors. Lower MAE and RMSE values indicate better temporal alignment between reconstructed and reference fiducial points, and thus greater preservation of clinically relevant waveform features.

For biomarker analysis, the QT interval was defined as the time from Q-wave onset to T-wave end, and the QRS duration was defined as the interval from QRS onset to QRS offset. Biomarker accuracy is also reported based on MAE and RMSE. 

% Detection performance is quantified via:
% \begin{equation}
%     \text{MAE} = \frac{1}{N}\sum_{i=1}^{N} |t_i - \hat{t}_i|
% \end{equation}
% \begin{equation}
%     \text{RMSE} = \sqrt{\frac{1}{N}\sum_{i=1}^{N} (t_i - \hat{t}_i)^2}
% \end{equation}
% where \(t_i\) denotes expert-annotated fiducial times and \(\hat{t}_i\) represents algorithm-detected points. Lower MAE/RMSE values indicate better temporal alignment between the reconstructed signal features and the ground truth annotations, reflecting the preservation of diagnostically relevant waveform morphology.

% Reconstructed signals were further analyzed for clinical biomarker accuracy using MAE and RMSE. The QT interval was measured from the onset of the Q-wave to the end of the T-wave, while the QRS duration was defined as QRS onset to QRS offset.

\section{Results}
\label{sec:result}

% \textcolor{red}{Let's add figures of the innovation signal synchronously averaged with the R-peaks. This will give us an objective idea of which parts of the ECG/PCG are easier/harder to reconstruct.}

All analyses in this section were performed using a segment window of $\Delta t = 0.5$\,s. Alternative window lengths (0.75\,s and 1\,s) were evaluated, but did not yield any improvement in model performance. Therefore, $\Delta t = 0.5$\,s was selected to ensure model simplicity and efficiency. To maintain clarity and focus, all reported results correspond to this window size. 

\subsection{Within-Subject Model Performance: Comparative Analysis}

Table~\ref{tab:whitin_metric} presents the evaluation metrics for the three modeling frameworks (LASSO, NN, and LSTM) under causal, anti-causal, and non-causal configurations for both ECG$\rightarrow$PCG and PCG$\rightarrow$ECG transformations in the within-subject scenario.

The results demonstrate that the LSTM model provides superior performance across all metrics and transformation directions. For example, in the non-causal setting, LSTM achieves a coherence of $0.39 \pm 0.16$, CC of $0.50 \pm 0.28$, and SNR of $1.9 \pm 2.1$~dB for ECG$\rightarrow$PCG, and even higher values for PCG$\rightarrow$ECG with coherence $0.76 \pm 0.08$, CC $0.80 \pm 0.19$, and SNR $7.6 \pm 6.1$~dB. This consistent outperformance highlights the LSTM's ability to capture complex temporal dependencies in multimodal cardiac signals.

A comparison of transformation directions reveals that PCG$\rightarrow$ECG is a less challenging problem than ECG$\rightarrow$PCG. For instance, across all models and configurations, the metrics for PCG$\rightarrow$ECG are substantially higher, indicating that reconstructing ECG from PCG is more effective, likely due to the greater information content and lower noise in the ECG signal.

\begin{figure*}[ht]
    \centering
    % First row
    \begin{subfigure}{0.24\textwidth}
        \includegraphics[trim={0cm 0cm 0cm 1.1cm},clip,width=\linewidth]{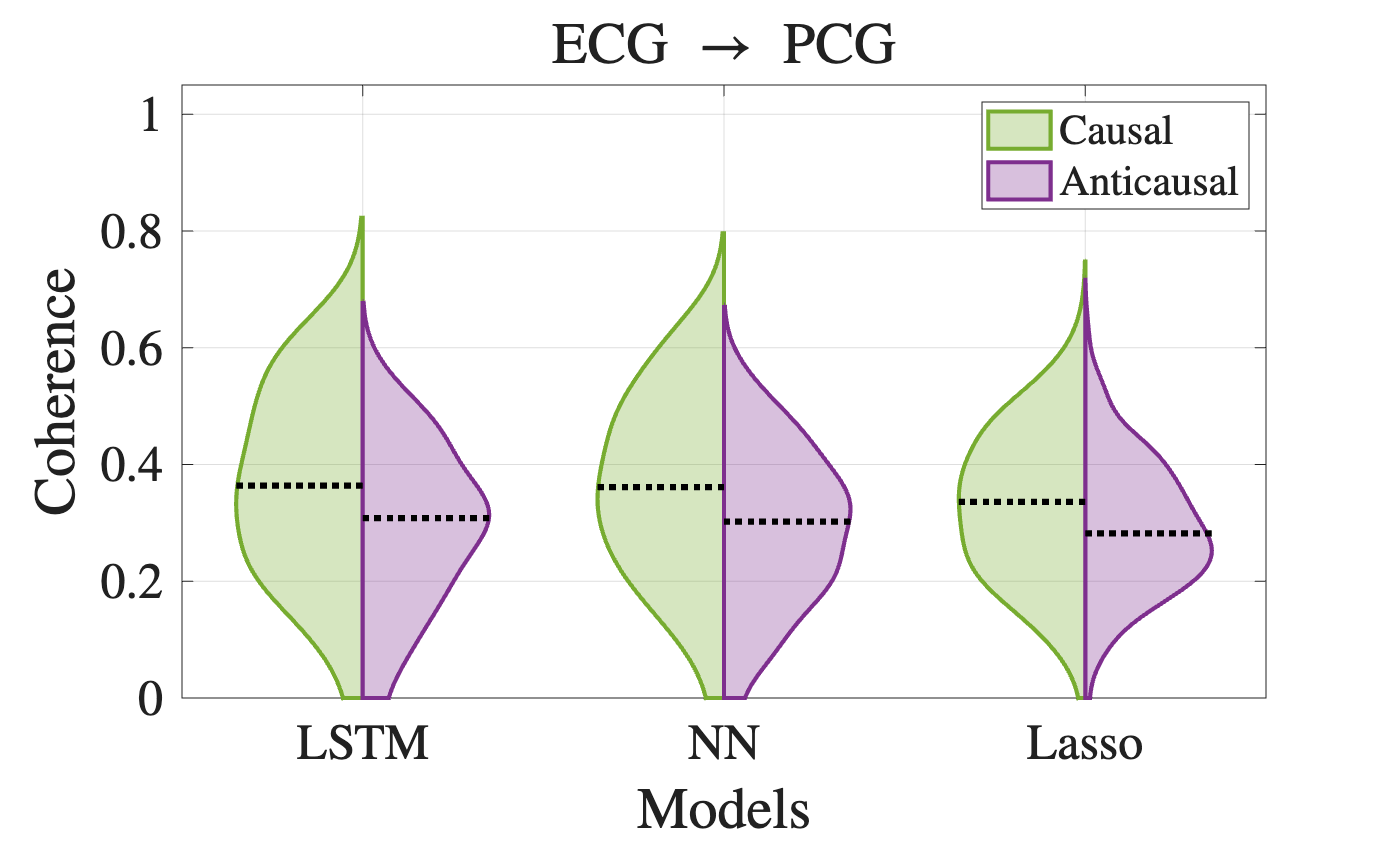}
        \caption{ECG$\rightarrow$PCG: Coherence}
    \end{subfigure}
    \hfill
    \begin{subfigure}{0.24\textwidth}
        \includegraphics[trim={0cm 0cm 0cm 1.1cm},clip,width=\linewidth]{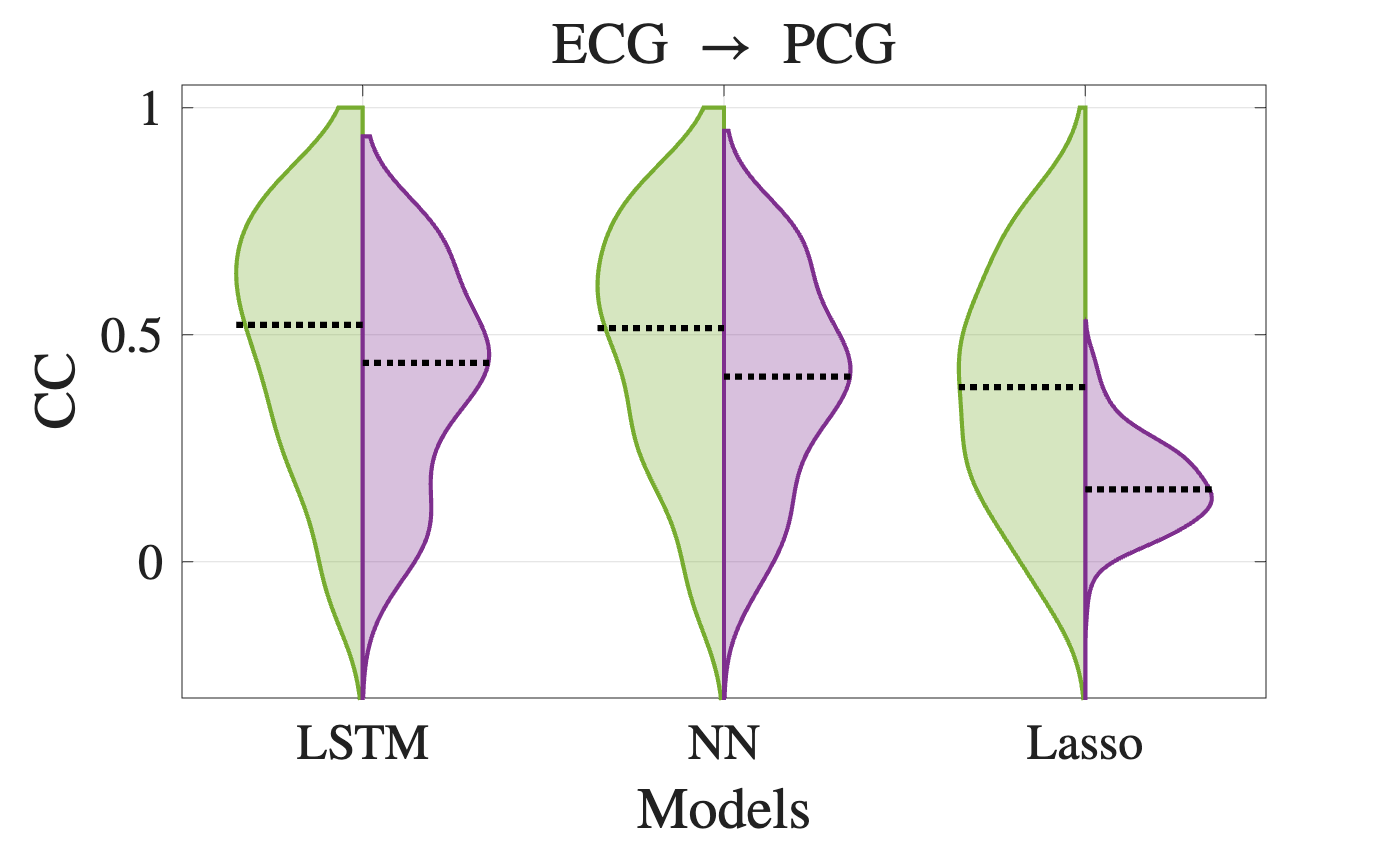}
        \caption{ECG$\rightarrow$PCG: CC}
    \end{subfigure}
    \hfill
    \begin{subfigure}{0.24\textwidth}
        \includegraphics[trim={0cm 0cm 0cm 1.1cm},clip,width=\linewidth]{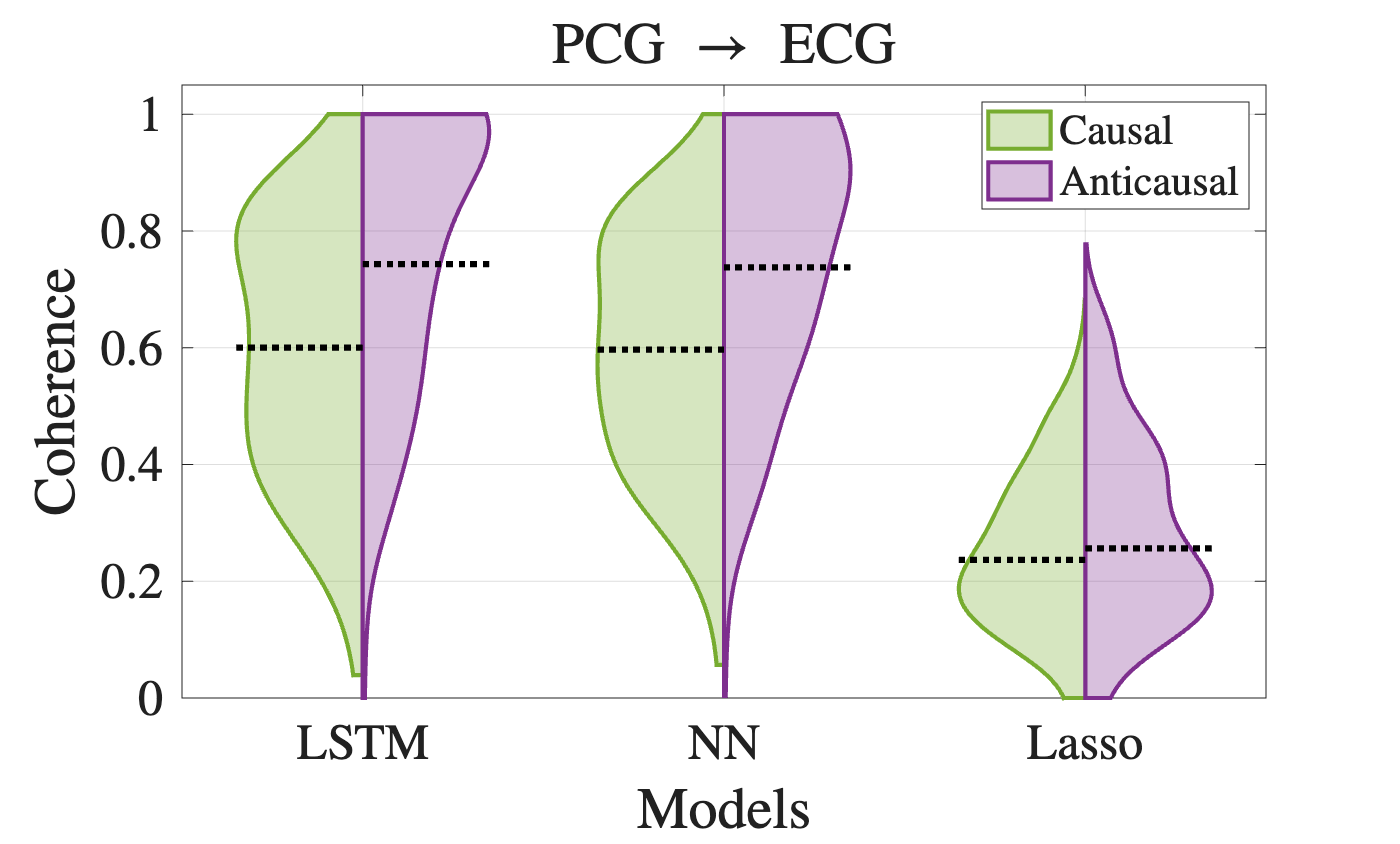}
        \caption{PCG$\rightarrow$ECG: Coherence}
    \end{subfigure}
    \hfill
    \begin{subfigure}{0.24\textwidth}
        \includegraphics[trim={0cm 0cm 0cm 1.1cm},clip,width=\linewidth]{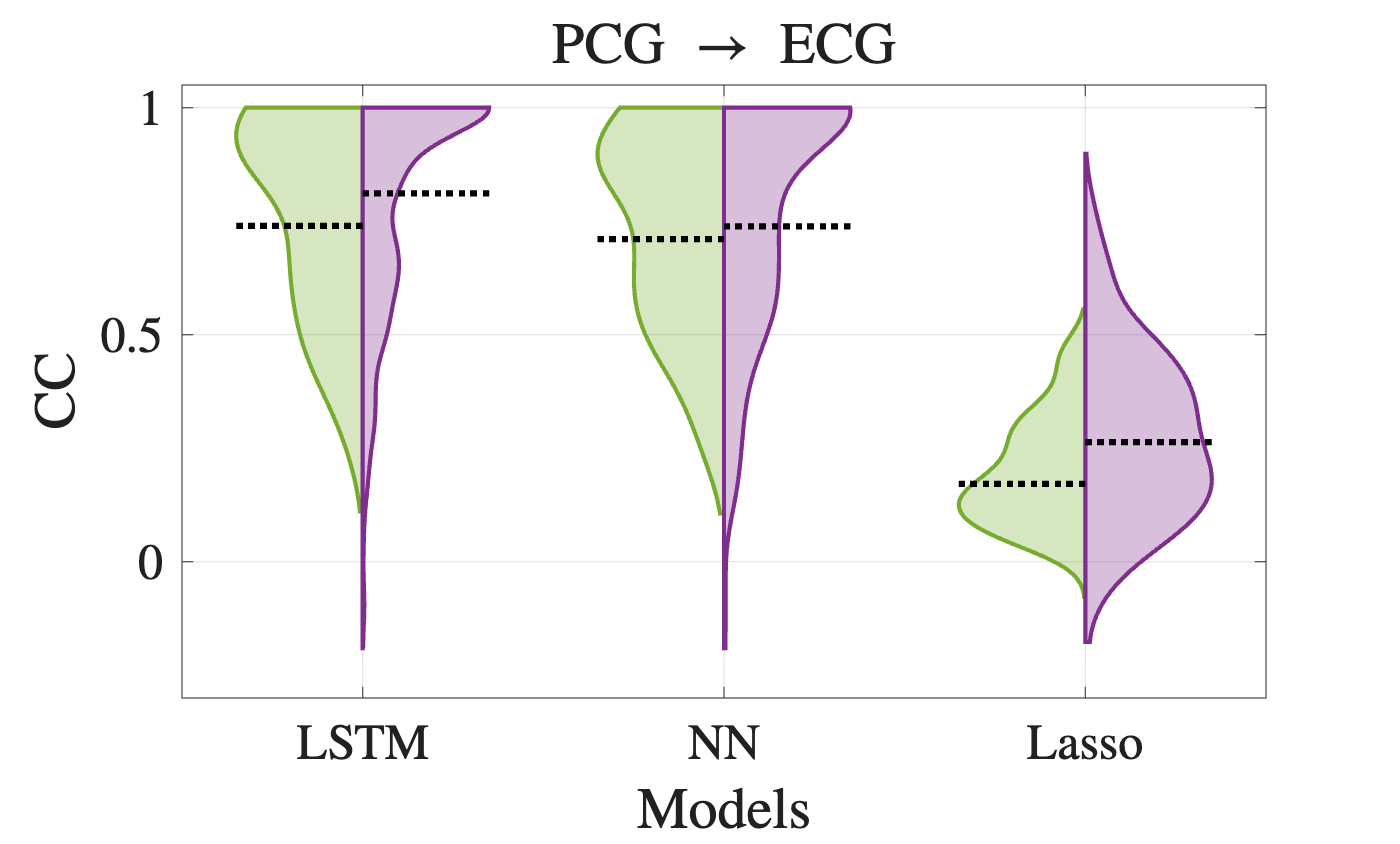}
        \caption{PCG$\rightarrow$ECG: CC }
    \end{subfigure}

    \vspace{0.5cm}

    \begin{subfigure}{0.24\textwidth}
        \includegraphics[trim={0cm 0cm 0cm 1.12cm},clip,width=\linewidth]{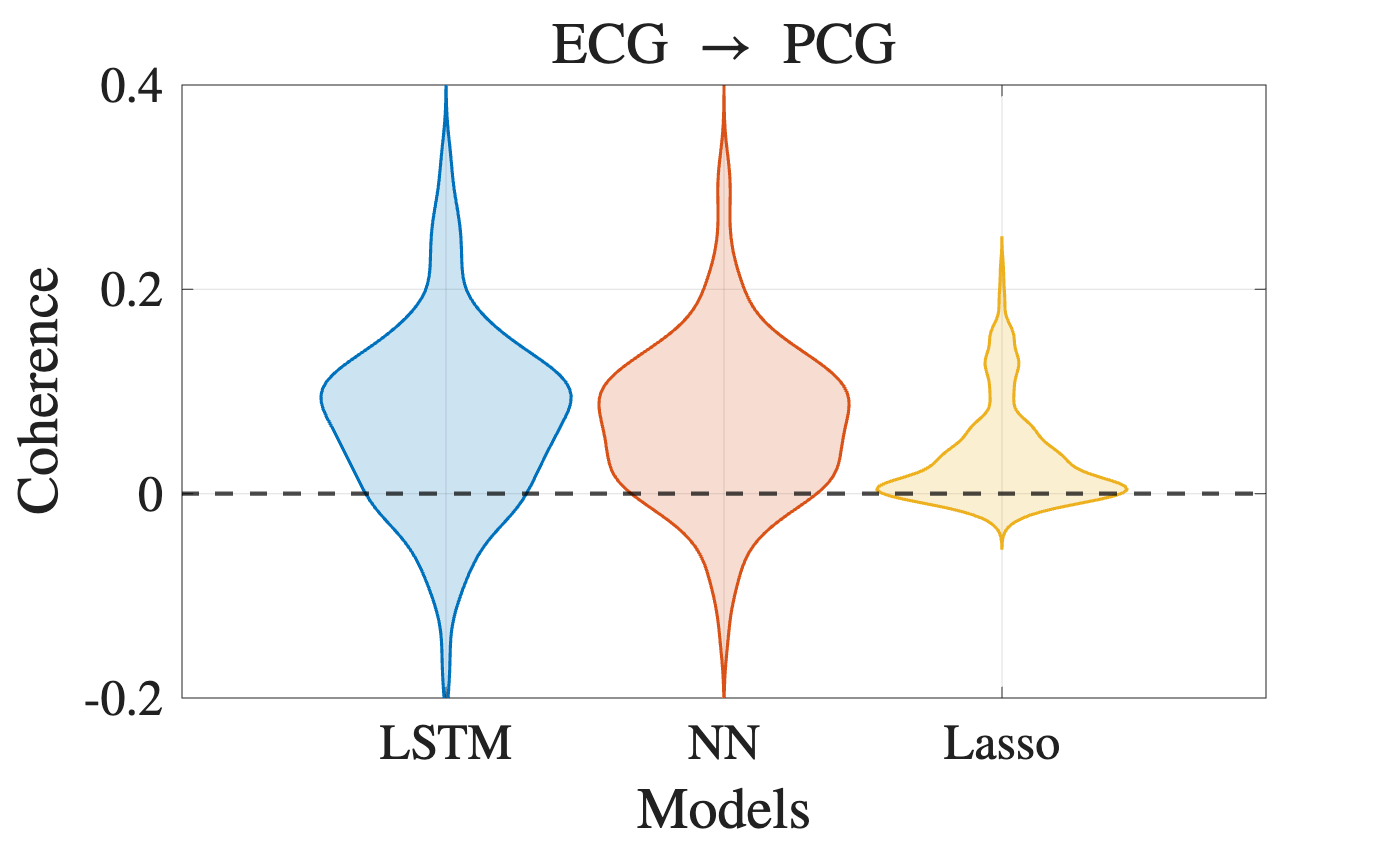}
        \caption{ ECG$\rightarrow$PCG: Coherence difference (causal $-$ anti-causal)}
    \end{subfigure}
    \hfill
    \begin{subfigure}{0.24\textwidth}
        \includegraphics[trim={0cm 0cm 0cm 1.12cm},clip,width=\linewidth]{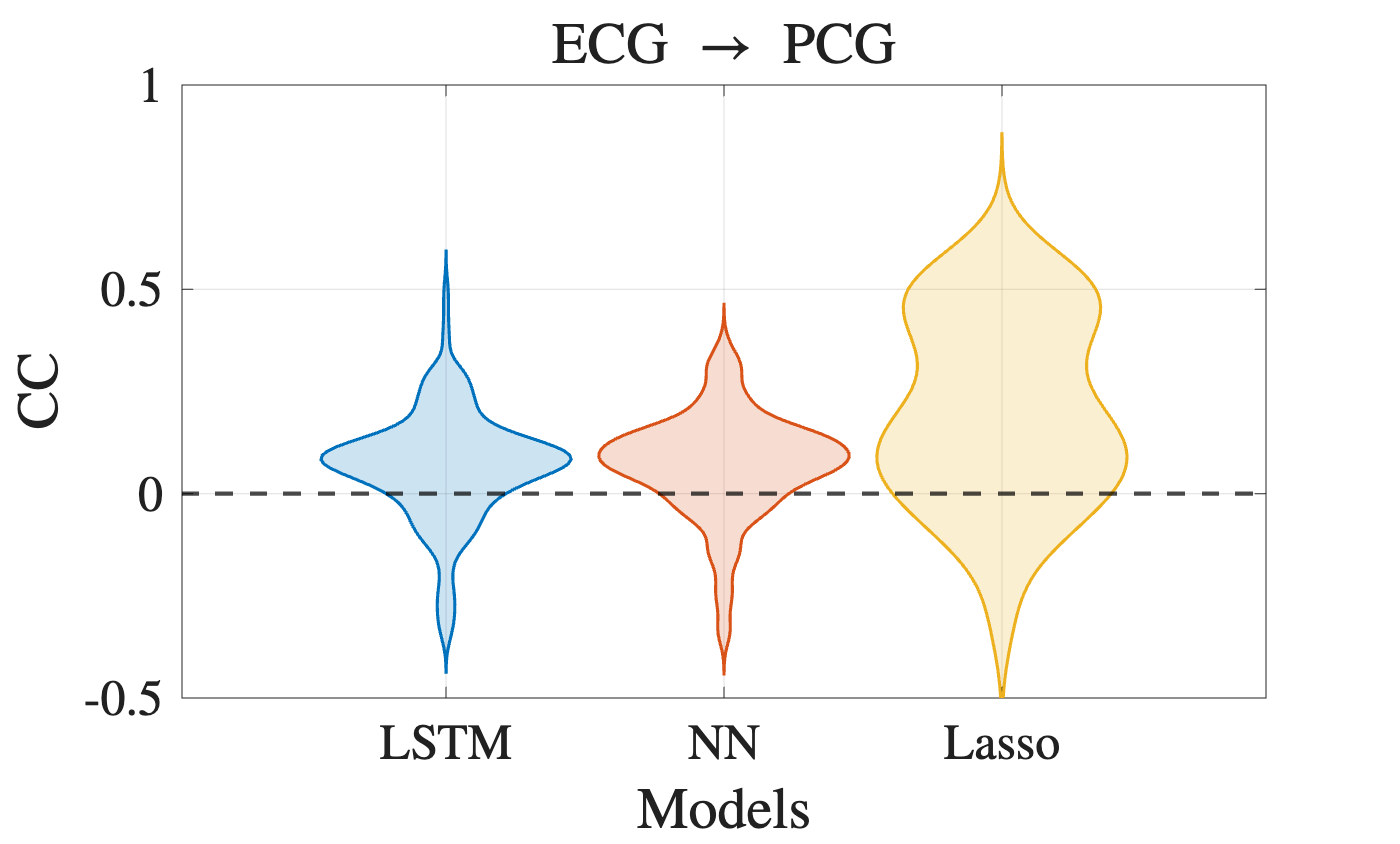}
        \caption{ECG$\rightarrow$PCG: CC difference (causal $-$ anti-causal)}
    \end{subfigure}
    \hfill
    \begin{subfigure}{0.24\textwidth}
        \includegraphics[trim={0cm 0cm 0cm 1.12cm},clip,width=\linewidth]{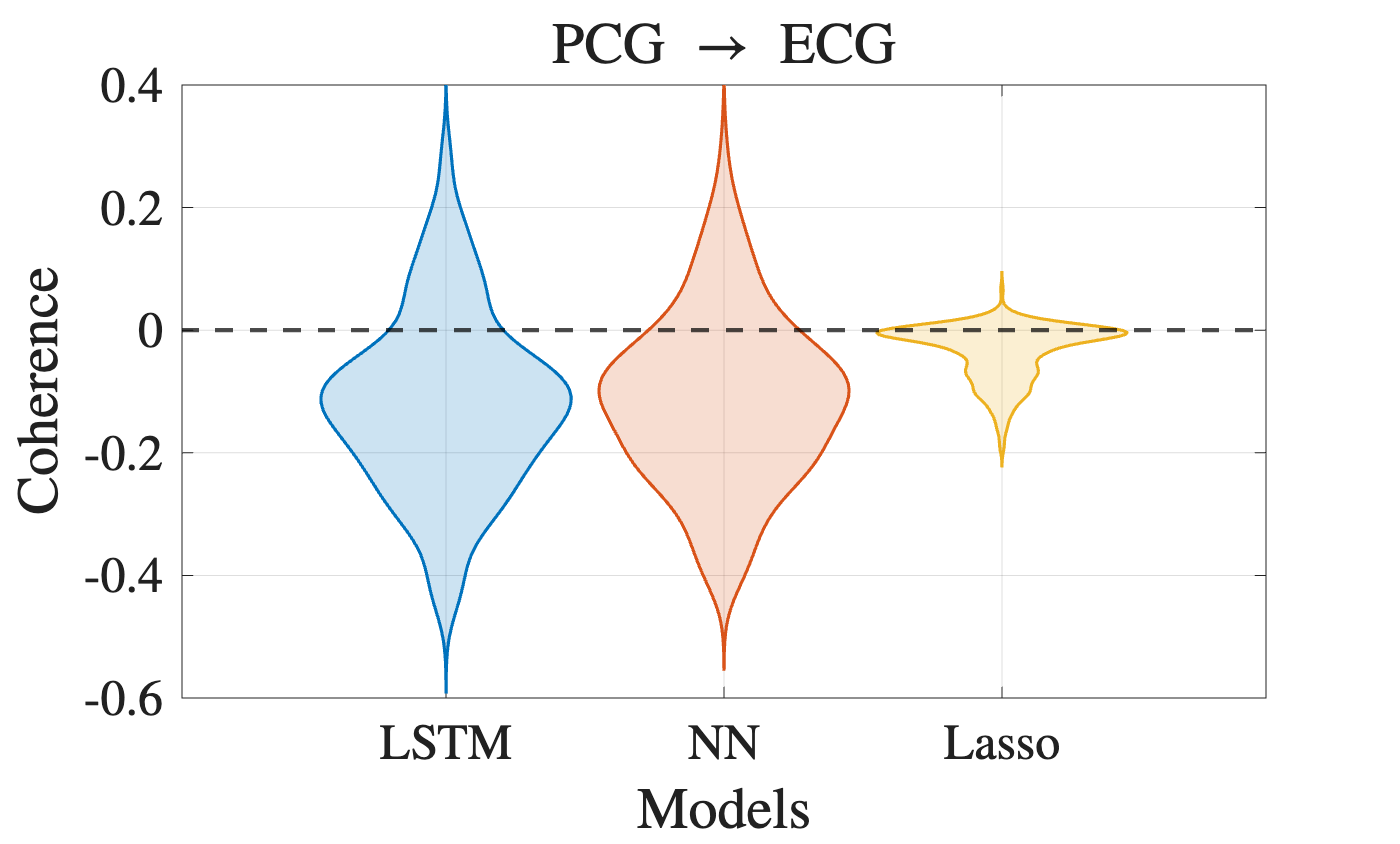}
        \caption{PCG$\rightarrow$ECG: Coherence difference (causal $-$ anti-causal)}
    \end{subfigure}
    \hfill
    \begin{subfigure}{0.24\textwidth}
        \includegraphics[trim={0cm 0cm 0cm 1.12cm},clip,width=\linewidth]{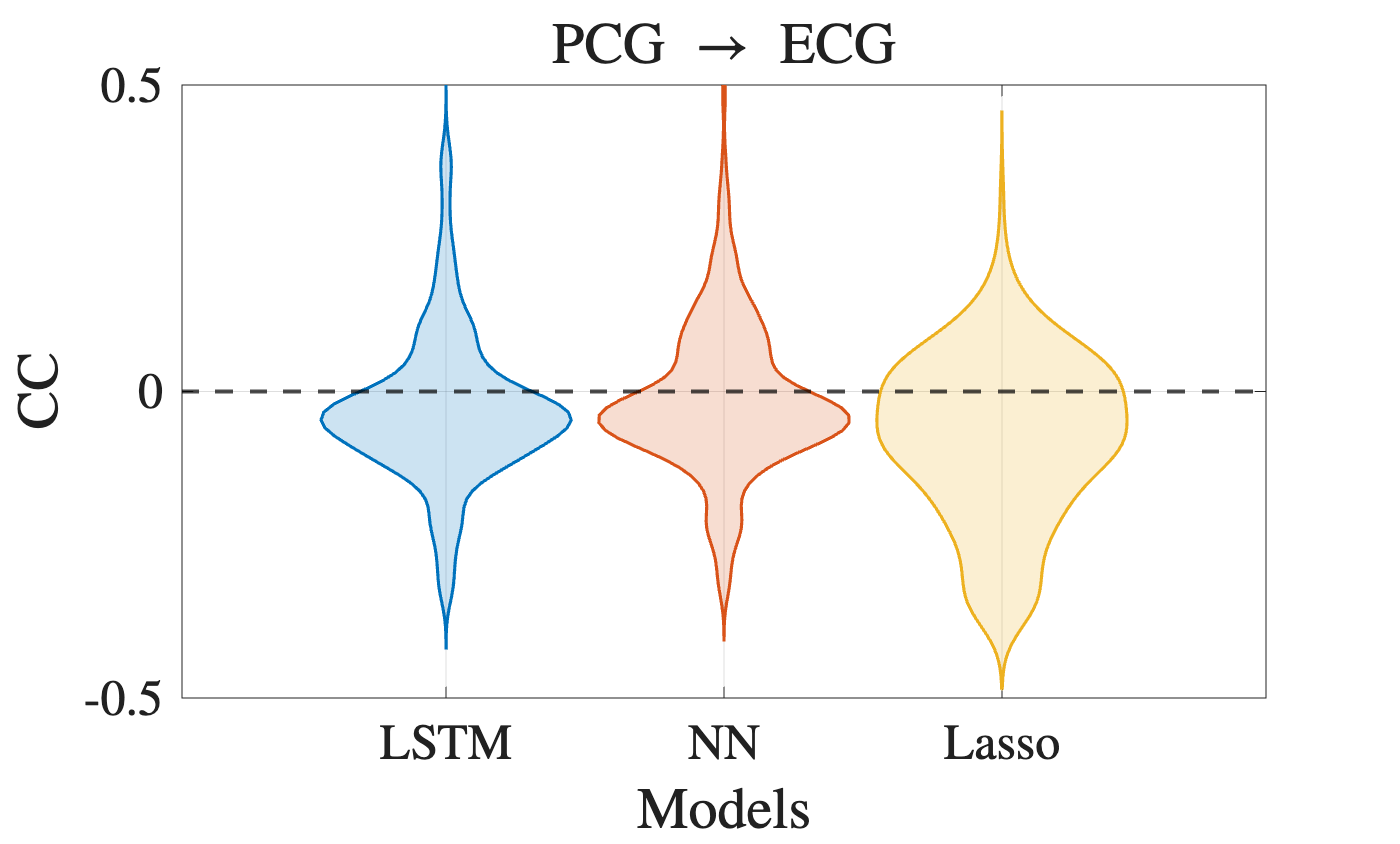}
        \caption{PCG$\rightarrow$ECG: CC difference (causal $-$ anti-causal)}
    \end{subfigure}
    \hfill
    
\caption{Violin plots illustrating evaluation metrics for the causal and anti-causal approaches in transforming ECG to PCG and vice versa. The first row displays the distributions of coherence and CC for each model and direction (causal vs. anti-causal), with the black dashed line indicating the average values. In contrast, the second row shows violin plots of the pairwise differences (causal minus anti-causal) for each metric and transformation. These visualizations offer insights into the directionality of information flow between modalities.}
    \label{fig:violin-plots}
\end{figure*}

Regarding causality, the results indicate that for ECG$\rightarrow$PCG, the causal configuration outperforms the anti-causal (e.g., LSTM coherence: $0.37$ causal vs. $0.30$ anti-causal), while for PCG$\rightarrow$ECG, the anti-causal configuration yields better performance (e.g., LSTM coherence: $0.71$ anti-causal vs. $0.59$ causal). This pattern is consistent with the physiological sequence of cardiac events, where electrical activity precedes mechanical response. The violin plots in Fig.~\ref{fig:violin-plots} further substantiate the findings regarding causality in multimodal cardiac signal transformation. In the first row, the distribution of coherence and CC for both causal and anti-causal configurations is shown for each model and transformation direction. For ECG$\rightarrow$PCG, the causal configuration consistently yields higher coherence and CC than the anti-causal configuration across all models (e.g., LSTM average coherence $\approx 0.37$ causal vs.\ $\approx 0.30$ anti-causal), indicating that predicting mechanical activity from preceding electrical activity aligns with physiological expectations. In contrast, for PCG$\rightarrow$ECG, the anti-causal configuration outperforms the causal one (e.g., LSTM average coherence $\approx 0.71$ anti-causal vs.\ $\approx 0.59$ causal), suggesting that reconstructing electrical activity from subsequent mechanical events is more effective, possibly due to the temporal structure of the cardiac cycle.
The second row of violin plots displays the distribution of the pairwise differences (causal minus anti-causal) for each metric. For ECG$\rightarrow$PCG, these differences are predominantly positive, confirming the advantage of the causal approach, while for PCG$\rightarrow$ECG, the differences are largely negative, reflecting the superiority of the anti-causal approach. This consistent trend across models and metrics highlights the importance of temporal directionality and supports the physiological basis for information flow between ECG and PCG signals.

In comparing linear and nonlinear models, coherence results show that LASSO performs comparably to NN and LSTM for ECG$\rightarrow$PCG (e.g., non-causal LASSO coherence: $0.33 \pm 0.13$ vs.\ LSTM $0.39 \pm 0.16$), but there is a pronounced gap for PCG$\rightarrow$ECG (e.g., non-causal LASSO coherence: $0.31 \pm 0.18$ vs.\ LSTM $0.76 \pm 0.08$), underscoring the importance of nonlinear modeling for accurate ECG reconstruction from PCG.

Finally, non-causal LSTM models consistently yield the highest performance for both ECG$\rightarrow$PCG and PCG$\rightarrow$ECG transformations, as evidenced by the highest coherence, CC, and SNR values in both tasks. In the following sections, we present a more detailed analysis of the non-causal LSTM results and their implications for multimodal cross-learning between ECG and PCG.

\begin{table}[tb]
\centering
\caption{Evaluation metrics for various models in causal, anti-causal, and non-causal approaches to ECG to PCG transformation and vice versa ($\hat{\text{PCG}}$ means ECG$\rightarrow$PCG and $\hat{\text{ECG}}$ means PCG$\rightarrow$ECG)}
\setlength{\extrarowheight}{2pt}
\begin{tabular}{|p{0.85cm}|p{0.6cm}|p{0.8cm}|c|c|c|}
\hline
\textbf{Analysis} & \textbf{State} & \textbf{Model} & \textbf{Coherence} & \textbf{CC} & \textbf{SNR}  \\ \hline
\multirow{9}{*}{$\hat{\text{PCG}}$} 
& \multirow{3}{0.5cm}{Causal} & LSTM    & 0.37 $\pm$ 0.16 & 0.47 $\pm$ 0.28 & 1.6 $\pm$ 2.0  \\ 
&                           & NN    & 0.36 $\pm$ 0.15 & 0.47 $\pm$ 0.28 & 1.6 $\pm$ 1.9  \\ 
&                           & LASSO  & 0.33 $\pm$ 0.13 & 0.38 $\pm$ 0.25 & 1.0 $\pm$ 1.4  \\ \cline{2-6}
& \multirow{3}{0.5cm}{Anti-causal} & LSTM  & 0.30 $\pm$ 0.13 & 0.41 $\pm$ 0.24 & 1.1 $\pm$ 1.3  \\ 
&                           & NN    & 0.30 $\pm$ 0.13 & 0.40 $\pm$ 0.23 & 1.0 $\pm$ 1.3  \\ 
&                           & LASSO & 0.29 $\pm$ 0.11 & 0.17 $\pm$ 0.10 & 0.2 $\pm$ 0.2  \\ \cline{2-6}
& \multirow{3}{0.5cm}{Non-causal} & LSTM  & 0.39 $\pm$ 0.16 & 0.50 $\pm$ 0.28 & 1.9 $\pm$ 2.1  \\ 
&                           & NN    & 0.37 $\pm$ 0.16 & 0.49 $\pm$ 0.27 & 1.8 $\pm$ 1.9  \\ 
&                           & LASSO & 0.33 $\pm$ 0.13 & 0.39 $\pm$ 0.26 & 1.1 $\pm$ 1.4  \\ \cline{2-6}
\noalign{\hrule height 1.2pt}
\multirow{9}{*}{$\hat{\text{ECG}}$} 
& \multirow{3}{0.5cm}{Causal} & LSTM    & 0.59 $\pm$ 0.21 & 0.71 $\pm$ 0.21 & 4.5 $\pm$ 3.6  \\ 
&                           & NN    & 0.59 $\pm$ 0.19 & 0.69 $\pm$ 0.21 & 4.0 $\pm$ 3.3  \\ 
&                           & LASSO  & 0.25 $\pm$ 0.13 & 0.20 $\pm$ 0.12 & 0.2 $\pm$ 0.3  \\ \cline{2-6}
& \multirow{3}{0.5cm}{Anti-causal} & LSTM  & 0.71 $\pm$ 0.22 & 0.74 $\pm$ 0.24 & 6.3 $\pm$ 5.7  \\ 
&                           & NN    & 0.71 $\pm$ 0.20 & 0.70 $\pm$ 0.25 & 5.2 $\pm$ 4.8  \\ 
&                           & LASSO & 0.29 $\pm$ 0.16 & 0.28 $\pm$ 0.18 & 0.5 $\pm$ 0.7  \\ \cline{2-6}
& \multirow{3}{0.5cm}{Non-causal} & LSTM  & 0.76 $\pm$ 0.8 & 0.80 $\pm$ 0.19 & 7.6 $\pm$ 6.1  \\ 
&                           & NN    & 0.75 $\pm$ 0.19 & 0.78 $\pm$ 0.20 & 6.5 $\pm$ 5.2  \\ 
&                           & LASSO & 0.31 $\pm$ 0.18 & 0.33 $\pm$ 0.20 & 0.7 $\pm$ 0.9  \\ \cline{1-6}
\end{tabular}
\raggedright
\label{tab:whitin_metric}
\end{table}

\subsection{Spectrum Characteristic Analysis}

\begin{figure}
\centering
  \begin{subfigure}[b]{0.9\columnwidth}
    \includegraphics[width=\linewidth]{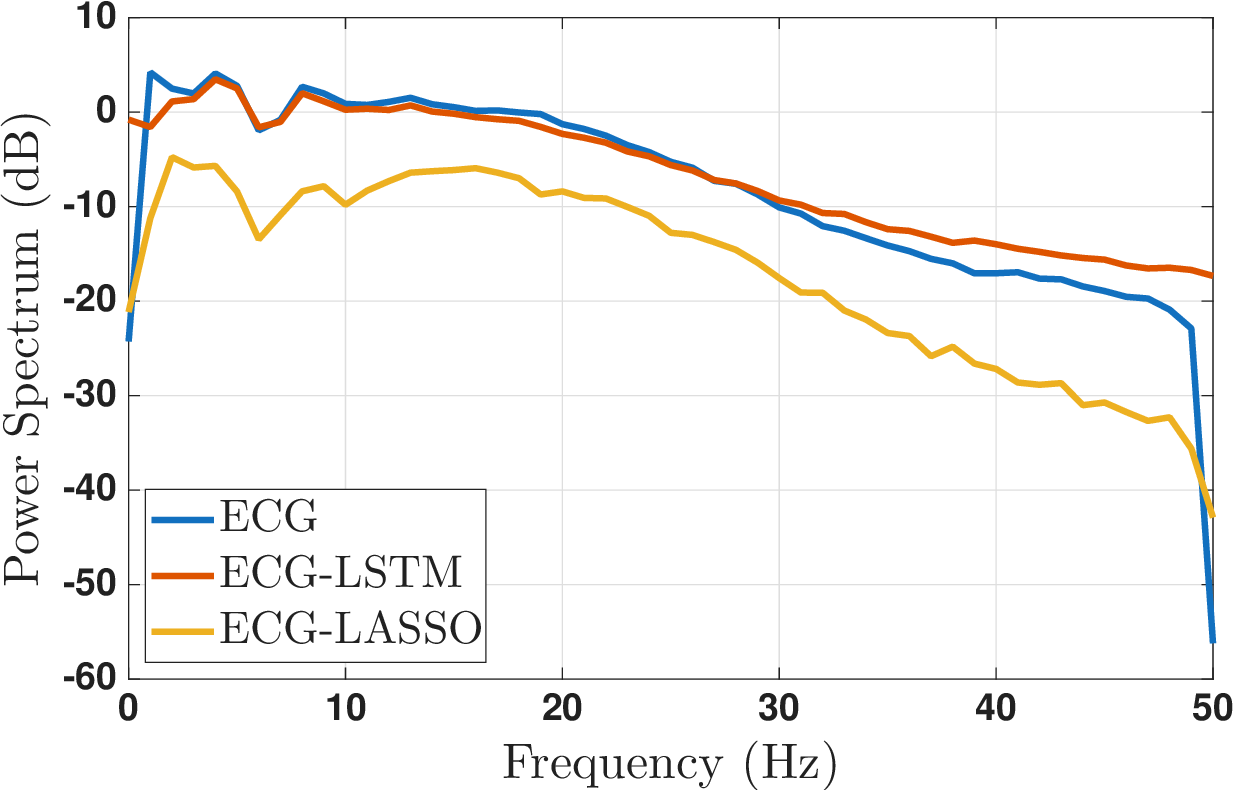}
    \caption{ECG power spectrum}
    \label{fig:viz_psecg}
  \end{subfigure}\\
  % \hfill %%
  \begin{subfigure}[b]{0.9\columnwidth}
    \includegraphics[width=\linewidth]{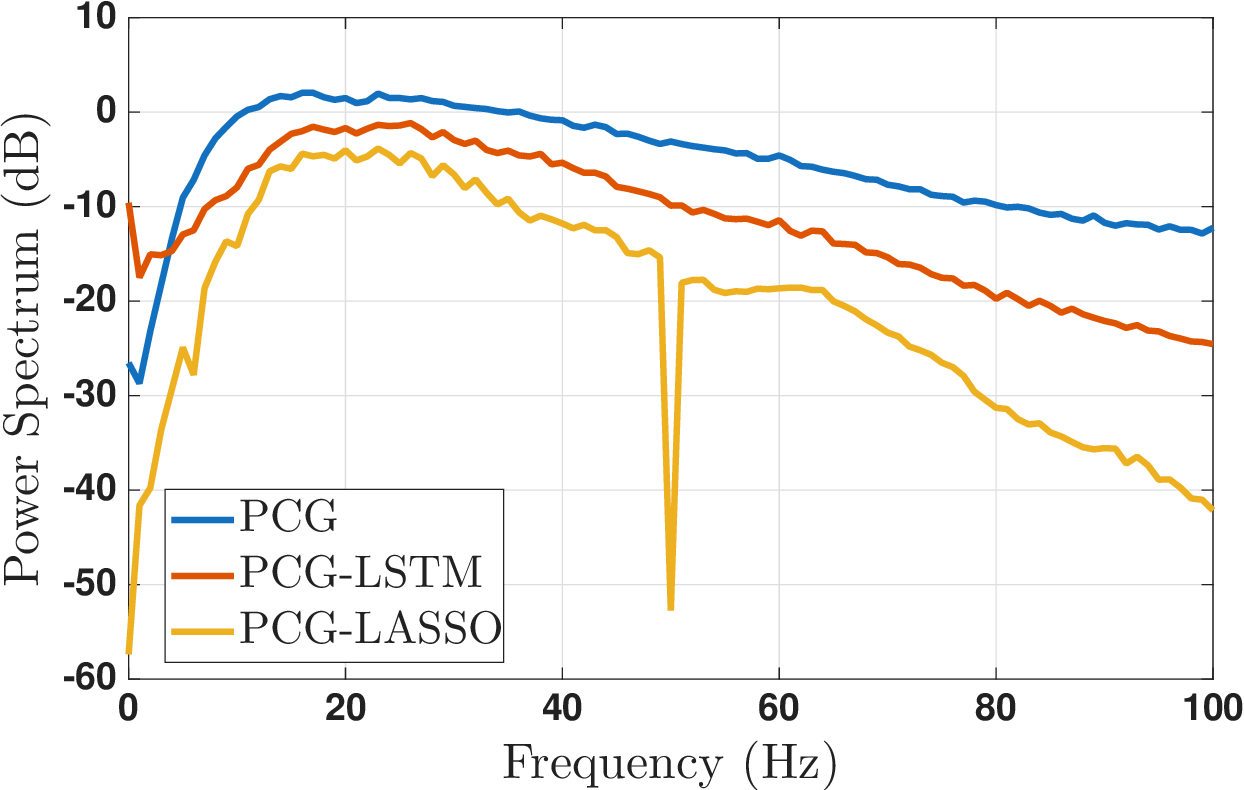}
    \caption{PCG power spectrum}
    \label{fig:viz_pspcg}
  \end{subfigure}
  \caption{Comparison of power spectrum for both original and reconstructed ECG and PCG signals by non-causal LSTM and LASSO in within-subject scenario. Subplots (a) and (b) show the power spectra, illustrating the spectral similarities and differences between the original and reconstructed waveforms for each modality. As expected, the nonlinear LSTM better learns the spectra in both cases.}
  \label{fig:viz_psd}
\end{figure}

To further elucidate the spectral properties of reconstructed signals, we compared the power spectra of original and reconstructed ECG and PCG waveforms using both the best-performing non-causal LSTM and the non-causal LASSO models (Fig.~\ref{fig:viz_psd}). This analysis provides insight into the frequency-domain fidelity of each reconstruction approach and highlights the limitations of linear versus nonlinear modeling frameworks.

For ECG reconstruction (Fig.~\ref{fig:viz_psecg}), the LSTM-reconstructed ECG (red line) closely follows the original ECG power spectrum (blue line) across most frequencies. Notably, the greatest discrepancies are observed below 4\,Hz, where the LSTM underestimates the power of very low-frequency components. This observation is consistent with the spectral characteristics of the original ECG shown in Fig.~\ref{fig:nspectra}, where unique low-frequency content is prominent. In contrast, the LASSO-reconstructed ECG (yellow line) exhibits a substantial reduction in spectral power across all frequencies, particularly in the low-frequency range, indicating the limited capacity of linear models to capture the full spectrum of ECG features. This aligns with the lower coherence and correlation metrics previously reported for LASSO in Table~\ref{tab:whitin_metric}.

For PCG reconstruction (Fig.~\ref{fig:viz_pspcg}), the LSTM-reconstructed PCG (red line) demonstrates a more pronounced deviation from the original PCG spectrum (blue line), especially at higher frequencies. The LSTM model captures the main spectral content below 40\,Hz but exhibits a faster decay compared to the original signal for frequencies higher than 40\,Hz.  Remarkably, due to the nonlinear modeling capacity of LSTM, the reconstructed PCG retains frequency components up to 80\,Hz. In comparison, the LASSO-reconstructed PCG (yellow line) is largely restricted to frequencies below 50\,Hz, with a notch valley at 50\,Hz due to the removal of the power-line interface in ECG using a notch filter, as mentioned in the preprocessing. This behavior aligns with the system-theoretic expectation that linear models (such as LASSO) cannot generate frequency content beyond what is present in the input ECG (see Fig.~\ref{fig:nspectra}), whereas nonlinear models like LSTM can interpolate or extrapolate additional frequency components.

% \begin{figure*}
% \centering
%   \begin{subfigure}[b]{0.75\columnwidth}
%     \includegraphics[width=\linewidth]{images/psdecg_lf.eps}
%     \caption{ECG power spectrum}
%     \label{fig:viz_psecg}
%   \end{subfigure}
%   % \hfill %%
%   \begin{subfigure}[b]{0.75\columnwidth}
%     \includegraphics[width=\linewidth]{images/psdpcg_lf.eps}
%     \caption{PCG power spectrum}
%     \label{fig:viz_pspcg}
%   \end{subfigure}\\
%   % \hfill %%
%   \begin{subfigure}[b]{0.75\columnwidth}
%     \includegraphics[width=\linewidth]{images/psdecgn.eps}
%     \caption{ECG spectral density}
%     \label{fig:viz_psecgn}
%   \end{subfigure}
%   % \hfill %%
%   \begin{subfigure}[b]{0.75\columnwidth}
%     \includegraphics[width=\linewidth]{images/psdpcgn.eps}
%     \caption{PCG spectral density}
%     \label{fig:viz_pspcg}
%   \end{subfigure}
%   \caption{Comparison of spectral characteristics for original and reconstructed ECG and PCG signals in the within-subject scenario using non-causal LSTM and LASSO models. Subplots (a) and (b) display the power spectra for ECG and PCG, respectively, while (c) and (d) present the corresponding power spectral densities.}
%   \label{fig:viz_psd4}
% \end{figure*}

\subsection{Performance Degradation Under Exercise Stress Conditions}

Table~\ref{tab:rest_excer} presents a comparative analysis of non-causal LSTM performance between resting (9 recordings) and exercise stress test conditions (19 recordings) for both transformation directions. The results demonstrate a consistent and substantial performance degradation during exercise across all evaluation metrics.

For ECG$\rightarrow$PCG transformation, the exercise condition shows marked deterioration compared to rest: coherence decreases from $0.54 \pm 0.11$ to $0.32 \pm 0.13$, correlation coefficient drops from $0.74 \pm 0.14$ to $0.39 \pm 0.26$, and SNR falls from $3.8 \pm 1.8$~dB to $0.9 \pm 1.4$~dB. Similarly, for PCG$\rightarrow$ECG transformation, exercise conditions result in reduced coherence ($0.93 \pm 0.09$ to $0.68 \pm 0.19$), correlation coefficient ($0.97 \pm 0.06$ to $0.72 \pm 0.18$), and SNR ($14.8 \pm 4.4$~dB to $4.2 \pm 3.1$~dB).

The larger performance gap observed for ECG$\rightarrow$PCG transformation (coherence drop of 0.22) and PCG$\rightarrow$ECG (coherence drop of 0.25) suggests that both transformation directions are similarly affected by exercise-induced complexity, though the absolute performance levels remain higher for the PCG$\rightarrow$ECG direction. These findings highlight the importance of considering physiological state when evaluating multimodal cardiac signal cross-learning algorithms and underscore the challenges of maintaining reconstruction fidelity under dynamic stress conditions.

This performance degradation can be attributed to two primary factors inherent to exercise stress testing. First, the presence of increased noise in both input and output signals during physical activity compromises signal quality and introduces artifacts that challenge the accuracy of cross-learning. Motion artifacts, muscle contractions, and respiratory variations contribute to elevated noise levels that are not present during resting conditions. Second, the dynamic physiological changes during exercise create more complex cardiac dynamics across varying heart rate ranges, from baseline through peak stress and recovery phases. These non-stationary conditions present a more challenging modeling scenario compared to the relatively stable hemodynamic state during rest.

\begin{table}[tb]
\centering
\caption{Performance comparison of non-causal LSTM for ECG-PCG bidirectional transformation between resting conditions (9 recordings) and exercise stress test conditions (19 recordings) using different evaluation metrics.}
\resizebox{\linewidth}{!}{
\setlength{\extrarowheight}{2pt}
\begin{tabular}{c|c|c|c|c|c}
\hline
\textbf{Analysis} & \textbf{Condition} & \textbf{Model} & \textbf{Coherence} & \textbf{CC} & \textbf{SNR}  \\ \hline \noalign{\hrule height 1.2pt}
\multirow{2}{*}{\textbf{ECG$\rightarrow$PCG}} 
& \multirow{1}{*}{Rest}          & LSTM    & 0.54 $\pm$ 0.11 & 0.74 $\pm$ 0.14 & 3.8 $\pm$ 1.8  \\ \cline{2-6}
& \multirow{1}{*}{Exercise} & LSTM    & 0.32 $\pm$ 0.13 & 0.39 $\pm$ 0.26 & 0.9 $\pm$ 1.4  \\ \hline 
\noalign{\hrule height 1.2pt}
\multirow{2}{*}{\textbf{PCG$\rightarrow$ECG}} 
& \multirow{1}{*}{Rest}          & LSTM     & 0.93 $\pm$ 0.09 & 0.97 $\pm$ 0.06 & 14.8 $\pm$ 4.4  \\ \cline{2-6}
& \multirow{1}{*}{Exercise} & LSTM     & 0.68 $\pm$ 0.19 & 0.72 $\pm$ 0.18 & 4.2 $\pm$ 3.1  \\ \cline{1-6}
\multicolumn{6}{c}{\phantom{\rule{0pt}{0.5em}}} \\ 
\end{tabular}
}
\raggedright
\label{tab:rest_excer}
\end{table}

\subsection{Generalization Performance: Within-Subject versus Cross-Subject Validation}

Table~\ref{tab:with_cross} compares the performance of non-causal LSTM models between within-subject and cross-subject validation scenarios for both transformation directions. The results reveal a substantial degradation in reconstruction performance when generalizing across subjects, highlighting the challenges of developing subject-independent multimodal cardiac signal models.

For ECG$\rightarrow$PCG transformation, cross-subject performance shows dramatic deterioration compared to within-subject results: coherence drops from $0.39 \pm 0.16$ to $0.11 \pm 0.08$, correlation coefficient decreases from $0.50 \pm 0.28$ to $0.03 \pm 0.10$, and SNR falls from $1.9 \pm 2.1$~dB to $-0.1 \pm 0.3$~dB. Similarly, for PCG$\rightarrow$ECG transformation, cross-subject validation yields reduced coherence ($0.76 \pm 0.20$ to $0.45 \pm 0.15$), correlation coefficient ($0.80 \pm 0.19$ to $0.43 \pm 0.26$), and SNR ($7.6 \pm 6.1$~dB to $1.0 \pm 1.3$~dB).

The pronounced performance degradation observed for both ECG$\rightarrow$PCG transformation (coherence drop of 0.28) and PCG$\rightarrow$ECG (coherence drop of 0.31) suggests that both directions are similarly affected by cross-subject variability, though the absolute performance levels remain higher for the PCG$\rightarrow$ECG direction. 

This performance degradation can be attributed to several physiological and technical factors. First, variations in ECG electrode placement and PCG stethoscope positioning across different recordings introduce systematic differences in signal morphology and amplitude that are not present in within-subject scenarios~\cite{clifford2006advanced}. These placement variations create additional uncertainty in the mapping between ECG and PCG features, as the spatial relationship between electrical and acoustic cardiac events becomes inconsistent. Second, individual differences in cardiac dynamics, including heart rate variability patterns, conduction system properties, and hemodynamic responses, contribute to subject-specific ECG-PCG relationships that cannot be easily generalized across populations.

\begin{table}[tb]
\centering
\caption{Performance comparison of non-causal LSTM models between within-subject and cross-subject validation scenarios for bidirectional ECG-PCG transformation using coherence, CC, and SNR metrics.}
\resizebox{\linewidth}{!}{
\setlength{\extrarowheight}{2pt}
\begin{tabular}{c|c|c|c|c|c}
\hline
\textbf{Analysis} & \textbf{Condition} & \textbf{Model} & \textbf{Coherence} & \textbf{CC} & \textbf{SNR}  \\ \hline
\multirow{2}{*}{\textbf{ECG$\rightarrow$PCG}} 
& \multirow{1}{*}{Within-subj.}  & LSTM  & 0.39 $\pm$ 0.16 & 0.50 $\pm$ 0.28 & 1.9 $\pm$ 2.1  \\   \cline{2-6}
& \multirow{1}{*}{Cross-subj.} & LSTM  & 0.11 $\pm$ 0.08 & 0.03 $\pm$ 0.10 & -0.1 $\pm$ 0.3  \\   \cline{2-6}
\noalign{\hrule height 1.2pt}
\multirow{2}{*}{\textbf{PCG$\rightarrow$ECG}} 
& \multirow{1}{*}{Within-subj.}  & LSTM  & 0.76 $\pm$ 0.20 & 0.80 $\pm$ 0.19 & 7.6 $\pm$ 6.1  \\  \cline{2-6}
& \multirow{1}{*}{Cross-subj.} & LSTM  & 0.45 $\pm$ 0.15 & 0.43 $\pm$ 0.26 & 1.0 $\pm$ 1.3  \\  \cline{1-6}
\end{tabular}
}
\raggedright
\label{tab:with_cross}
\end{table}

\begin{table}[tb]
\centering
\caption{Performance comparison of non-causal LSTM models for instantaneous amplitude reconstruction between within-subject and cross-subject validation scenarios, demonstrating improved cross-subject generalization compared to raw signal reconstruction.}
\resizebox{\linewidth}{!}{
\setlength{\extrarowheight}{2pt}
\begin{tabular}{c|c|c|c|c|c}
\hline
\textbf{Analysis} & \textbf{Condition} & \textbf{Model} & \textbf{Coherence} & \textbf{CC} & \textbf{SNR}   \\ \hline
\multirow{2}{*}{\textbf{ECG$\rightarrow$PCG}} 
& \multirow{1}{*}{Within-subj.}  & LSTM  & 0.86 $\pm$ 0.11 & 0.74 $\pm$ 0.18 & 6.7 $\pm$ 2.5  \\   \cline{2-6}
& \multirow{1}{*}{Cross-subj.} & LSTM  & 0.84 $\pm$ 0.10 & 0.63 $\pm$ 0.14 & 4.5 $\pm$ 1.0  \\   \cline{2-6}
\noalign{\hrule height 1.2pt}
\multirow{2}{*}{\textbf{PCG$\rightarrow$ECG}} 
& \multirow{1}{*}{Within-subj.}  & LSTM  & 0.91 $\pm$ 0.11 & 0.80 $\pm$ 0.21 & 12.1 $\pm$ 6.0  \\  \cline{2-6}
& \multirow{1}{*}{Cross-subj.} & LSTM  & 0.85 $\pm$ 0.10 & 0.60 $\pm$ 0.16 & 6.2 $\pm$ 1.7  \\  \cline{1-6}
\end{tabular}
}
\raggedright
\label{tab:with_cross_abs}
\end{table}

To mitigate the aggregate effects of all sources of inter-subject variation, which may include anatomical and technical factors in cross-subject scenarios, this study investigated instantaneous amplitude (envelope) estimation using the analytic signal representation. The instantaneous amplitude is derived from the Hilbert transform, which decomposes the signal into its envelope and phase components according to:

\begin{equation}
x(t) = A(t) e^{j\phi(t)}
\end{equation}
where $A(t)$ represents the instantaneous amplitude (envelope) and $\phi(t)$ denotes the instantaneous phase. While the analytic signal is a mathematical construct and does not directly address electrode placement or anatomical variability, using the envelope as a modeling target enables a more robust estimation of signal intensity across subjects, as it summarizes the overall energy profile independently of phase alignment. By focusing on the instantaneous amplitude $A_{\text{ECG}}(t)$ and $A_{\text{PCG}}(t)$ as reconstruction targets rather than the original ECG and PCG signals, the model becomes less sensitive to phase misalignments and morphological variations that arise from inter-subject differences in electrode positioning and anatomical variability.

Table~\ref{tab:with_cross_abs} presents the performance comparison for instantaneous amplitude reconstruction between within-subject and cross-subject validation scenarios. The results demonstrate substantial improvements in cross-subject generalization compared to the raw signal reconstruction shown in Table~\ref{tab:with_cross}. For ECG$\rightarrow$PCG amplitude transformation, the cross-subject coherence reaches $0.84 \pm 0.10$ compared to only $0.11 \pm 0.08$ for raw signals (Table~\ref{tab:with_cross}), representing a significant improvement of 0.73. Similarly, for PCG$\rightarrow$ECG amplitude transformation, cross-subject coherence achieves $0.85 \pm 0.10$ versus $0.45 \pm 0.15$ for raw signals, showing an improvement of 0.40.

Notably, the performance gap between within-subject and cross-subject scenarios is substantially reduced when using amplitude targets. For ECG$\rightarrow$PCG, the coherence difference decreases from 0.28 (raw signals) to only 0.02 (amplitude), while for PCG$\rightarrow$ECG, the gap reduces from 0.31 to 0.06. This demonstrates that amplitude-based reconstruction effectively addresses the primary sources of cross-subject variability, making the approach more suitable for clinical applications where subject-independent models are essential.

\subsection{Evaluation of ECG Fiducial Point Detection Accuracy}

The EPHNOGRAM dataset provides expert-annotated fiducial points for QRSon, R-peak, QRSoff, Ton, T-peak, and Toff across 23 exercise and 9 rest recordings, with approximately 180 heartbeats per recording, totaling 4,966 annotated heartbeats for the current study.

Table~\ref{tab:fid-ecg} summarizes the detection accuracy of ECG fiducial points under five different reconstruction schemes: original ECG, within-subject reconstructed ECG (WS), within-subject reconstructed ECG envelope (WS-ENV), cross-subject reconstructed ECG (CS), and cross-subject reconstructed ECG envelope (CS-ENV). For each ground truth fiducial point, detected points within a 200-ms window were counted as detected; otherwise, they were considered misses.

Table~\ref{tab:fid-ecg} demonstrates that the original ECG signals demonstrate the lowest MAE and RMSE across all fiducial points, with MAE values of 9.1 ms for QRSon, 6.1 ms for R-peak, and 17.2 ms for Toff, and sensitivities above 98\%. Within-subject reconstructed ECG (WS) also achieves high accuracy, with MAE values of 12.7 ms for QRSon, 10.8 ms for R-peak, and 22.2 ms for Toff, and sensitivities around 91\%. Envelope-based within-subject reconstruction (WS-ENV) shows slightly increased errors, particularly for Toff (MAE: 36.9 ms), indicating greater distortion in the detection of T-wave offset.

In the cross-subject scenario, CS-ENV outperforms CS in both sensitivity and error metrics. CS-ENV achieves an average sensitivity improvement of approximately 5\% over CS (e.g., 86.0\% vs.\ 79.6\% for Toff), and lower RMSE for most fiducial points except Toff. For example, CS-ENV achieves a sensitivity of 88.3\% for R-peak and 85.9\% for both QRSon and Toff, compared to 85.0\% and 79.3\% for CS, respectively. 

Compared to the original ECG, CS-ENV exhibits increased MAE for R-peak detection (from 6.1 ms to 19.7 ms), QRSon (from 9.1 ms to 18.2 ms), and Toff (from 17.2 ms to 41 ms), and a decrease in sensitivity from 99.0\% to 88.3\% for R-peak. Despite these increases in error, CS-ENV provides a practical solution for estimating ECG fiducial points from unseen PCGs in cross-subject applications, thereby balancing generalizability and detection accuracy.

\begin{table}[tb]
\centering
\caption{Fiducial point detection errors (MAE and RMSE, ms), sensitivity (\%), and number of detected points (N) for QRS and T-wave across five signal types: original ECG, within-subject reconstructed ECG (WS), within-subject reconstructed ECG envelope (WS-ENV), cross-subject reconstructed ECG (CS), and cross-subject reconstructed ECG envelope (CS-ENV). Total expert-annotated beats: 4,966.}
\resizebox{0.5\textwidth}{!}{
\begin{tabular}{|c|c|c|c|c|c|c|c|}
\hline
            Metrics & Type &  {QRSon} & {R-peak} & {QRSoff} & {Ton} & {T-peak} & {Toff}  \\ \hline
            \multirow{5}{*}{\textbf{MAE}} 
            &{ ECG }            &  9.1 & 6.1 & 8.8 & 12.1 & 5.7 & 17.2 \\ 
            &{ WS}              &  12.7 & 10.8 & 13.2 & 15.8 & 13.2 & 22.2 \\ 
            &{ WS-ENV}          &  13.2 & 12.2 & 19.8 & 18.5 & 17.4 & 36.9 \\
            &{ CS}              &  17.7 & 24.6 & 18.4 & 24.4 & 30.4 & 34.4 \\ 
            &{ CS-ENV}          &  18.2 & 19.7 & 17.0 & 22.7 & 30.3 & 41 \\  \hline
\multirow{5}{*}{\textbf{RMSE}} 
            &{  ECG }          &  21.0 & 20.2 & 20.0 & 22.4 & 22.1 & 34.9 \\ 
            &{ WS}             &  28.2 & 28.2 & 27.6 & 28.9 & 29.8 & 40.6 \\ 
            &{ WS-ENV}         &  26.3 & 27.5 & 43.4 & 32.4 & 31.5 & 51.6 \\
            &{ CS}             &  32.0 & 40.5 & 32.7 & 40.4 & 47.6 & 53.3 \\ 
            &{ CS-ENV}         & 31.2 & 33.8 & 31.1 & 36.9 & 44.9 & 56.9 \\ \hline
            
\multirow{5}{*}{\textbf{N}} 
            &{ ECG }           &  4890 & 4916 & 4888 & 4890 & 4891 & 4889 \\ 
            &{ WS}             &  4513 & 4603 & 4502 & 4499 & 4502 & 4498 \\ 
            &{ WS-ENV}         &  4527 & 4599 & 4525 & 4368 & 4355 & 4378 \\
            &{ CS}             &  3940 & 4222 & 3931 & 3955 & 3950 & 3953 \\ 
            &{ CS-ENV}         &  4264 & 4387 & 4262 & 4266 & 4268 & 4270 \\ \hline
\multirow{5}{*}{\textbf{Sen \%}} 
            &{ ECG }           &  98.5 & 99.0 & 98.4 & 98.5 & 98.5 & 98.4 \\ 
            &{ WS}             &  90.9 & 92.7 & 90.7 & 90.6 & 90.7 & 90.6 \\ 
            &{ WS-ENV}         &  91.2 & 92.6 & 91.1 & 88.0 & 87.7 & 88.2 \\
            &{ CS}             &  79.3 & 85.0 & 79.2 & 79.6 & 79.5 & 79.6 \\ 
            &{ CS-ENV}         &  85.9 & 88.3 & 85.8 & 85.9 & 85.9 & 86.0 \\ \hline
\end{tabular}
}
\label{tab:fid-ecg}
\end{table}

We further evaluated the accuracy of key ECG biomarkers—QT interval and QRS duration—extracted from reconstructed signals. For the cross-subject envelope-based approach (CS-ENV), the MAE for QT interval estimation was 38.8 ms, compared to 17.6 ms for the original ECG, and the RMSE was 55.2 ms versus 32.7 ms, respectively. For QRS duration, CS-ENV achieved an MAE of 13.9 ms and an RMSE of 18.8 ms, compared to 9.1 ms and 14.2 ms for the original ECG. While errors were higher for CS-ENV than for direct ECG annotation, these results demonstrate that clinically relevant intervals can be estimated from PCG-driven ECG reconstructions with reasonable accuracy, supporting the practical applicability of the proposed cross-subject modeling framework for biomarker extraction.

% QT interval
% Org ECG 17.6 while CS-ENV 38.8 in MAE  
% Org ECG 32.7 while CS-ENV 55.2 in RMSE  

% QRS duration
% Org ECG 9.1 while CS-ENV 13.9 in MAE  
% Org ECG 14.2 while CS-ENV 18.8 in RMSE  

\section{Discussion}
\label{sec:discussion}
This study aimed to systematically investigate the information transfer and shared versus exclusive characteristics between ECG and PCG signals, utilizing the EPHNOGRAM dataset and a suite of linear and nonlinear machine learning frameworks. By employing causal, anti-causal, and non-causal models, and evaluating within- and cross-subject generalization, we provide a comprehensive perspective on the potential and limitations of multimodal cardiac waveform learning.

The results demonstrate that nonlinear models, particularly non-causal LSTM architectures, are highly effective for reconstructing one modality from the other, outperforming linear approaches across all metrics. This finding aligns with recent advances in multimodal cardiac signal analysis, where the complexity and nonlinearity of electromechanical coupling are better captured by deep learning frameworks~\cite{hettiarachchi2021novel, han2023multimodal, ajitkumar2021heart}. The superiority of the LSTM model is evident not only in waveform recovery metrics but also in its ability to preserve the spectral and temporal features necessary for downstream biomarker extraction.

Spectral analyses reinforce the superiority of the non-causal LSTM model for reconstructing both ECG and PCG waveforms, while also illustrating the inherent challenges in recovering modality-specific frequency components, particularly the low-frequency content unique to ECG and the high-frequency content unique to PCG. The results show that linear models, such as LASSO, are inadequate for the PCG$\rightarrow$ECG transformation because non-overlapping frequencies in the spectra require a nonlinear model for effective mapping between two modalities.

A key insight from this work is the asymmetry in transformation difficulty: reconstructing ECG from PCG is generally more feasible and accurate than reconstructing PCG from ECG. This observation is consistent with the physiological role of ECG as a direct measure of cardiac electrical activity, which underpins the mechanical events captured by PCG~\cite{castro2015analysis}. Causality analysis further reveals that the optimal temporal direction for model input depends on the transformation task, reflecting the underlying electromechanical sequence of the cardiac cycle.

The physiological state has a pronounced effect on model performance. During exercise stress testing, increased noise, motion artifacts, and rapidly changing heart dynamics substantially degrade the accuracy of both ECG$\rightarrow$PCG and PCG$\rightarrow$ECG reconstruction. These findings underscore the importance of robust preprocessing and adaptive modeling when deploying multimodal learning in ambulatory or high-variability settings.

Generalization across subjects remains a significant challenge, as demonstrated by the marked performance drop in cross-subject validation. This limitation is primarily attributed to inter-individual differences in cardiac anatomy and dynamics, as well as variability in sensor placement~\cite{scholzel2016can,clifford2006advanced}. However, we show that shifting the modeling target from raw waveforms to instantaneous amplitude (envelope) features dramatically improves cross-subject performance. Envelope-based models exhibit high coherence and correlation, with only minor reductions compared to within-subject results, indicating that amplitude features are more robust to phase and morphological variability.

Furthermore, the study demonstrates that envelope-based cross-subject models can achieve practical accuracy not only for ECG fiducial point detection (such as QRSon, R-peak, and Toff) but also for key biomarker estimation, including QT interval and QRS duration, using PCG alone. While the precision does not match that of the original ECG, the achieved mean absolute errors and sensitivities support the feasibility of PCG-driven ECG biomarker estimation in real-world and resource-limited settings, thereby broadening the clinical utility of multimodal cardiac monitoring.

Taken together, these findings highlight the value of integrating ECG and PCG for comprehensive cardiac assessment. Multimodal learning not only leverages the complementary strengths of each modality but also addresses the limitations inherent in single-signal approaches~\cite{hettiarachchi2021novel, han2023multimodal, ajitkumar2021heart}. The demonstrated effectiveness of nonlinear models and amplitude-based targets for cross-subject generalization paves the way for more robust, accessible, and patient-specific cardiac monitoring solutions. Future research should investigate advanced architectures, such as transformers and generative adversarial networks, and further explore domain adaptation strategies to enhance generalizability and clinical applicability.

\section{Conclusion}
\label{sec:conclusion}

This work provides a unified framework for understanding and modeling the shared and exclusive information between ECG and PCG signals through data-driven, multimodal learning. Nonlinear models, especially non-causal LSTM architectures, consistently yield superior performance for both ECG$\rightarrow$PCG and PCG$\rightarrow$ECG transformations, with the latter direction proving more tractable. The study demonstrates that physiological state and inter-subject variability are critical factors influencing model accuracy; however, envelope-based modeling substantially mitigates these effects, enabling robust cross-subject generalization.

By showing that clinically relevant ECG biomarkers can be estimated from PCG in a generalizable manner, this study advances the potential for multimodal cardiac monitoring in both clinical and resource-constrained environments. These results support the continued development of multimodal, machine learning-based approaches for cardiac signal analysis, with an emphasis on robust, generalizable, and interpretable models for future clinical translation.

\section*{Acknowledgment}
This research was supported by the American Heart Association Innovative Project Award 23IPA1054351, on ``developing multimodal cardiac biomarkers for cardiovascular-related health assessment.''

\bibliographystyle{IEEEtran}
\bibliography{References}
%////////////////////////////////
\end{document}